\documentclass[smallextended]{svjour3}       
\smartqed

\usepackage{amssymb}
\usepackage{latexsym}
\usepackage{lineno,hyperref}
\usepackage{amsmath}
\usepackage{amsfonts} 
\usepackage{graphicx}
\usepackage{epsfig}
\usepackage{subcaption}
\usepackage{mathtools}
\usepackage{caption}
\usepackage{eso-pic}

\begin{document}
\AddToShipoutPictureBG*{%
  \AtPageUpperLeft{%
    \hspace{\paperwidth}%
    \raisebox{-\baselineskip}{%
      \makebox[0pt][r]{\textit{Paper accepted: Applied Intelligence (Springer), expected in 2020.}
}}}}%
%

\title{Improving cluster recovery with feature rescaling factors}

\author{Renato Cordeiro de Amorim \and
        Vladimir Makarenkov 
}

\authorrunning{R.C. de Amorim \and V. Makarenkov} 

\institute{R.C. de Amorim \at
			School of Computer Science and Electronic Engineering, University of Essex, Wivenhoe Park, CO4 3SQ, UK.\\
              Tel.: +44 (0) 1206 872895\\
              \email{r.amorim@essex.ac.uk}           
           \and
           V. Makarenkov \at
              D\'epartement  d'informatique,  Universit \'e du  Qu\'ebec \`a  Montr\'eal,  C.P.  8888  succ.  Centre-Ville,  Montreal(QC) H3C 3P8 Canada.\\
              \email{makarenkov.vladimir@uqam.ca}           
}

\date{Received: 01/07/2020 Accepted: 01/12/2020}

\maketitle

\begin{abstract}
The data preprocessing stage is crucial in clustering. Features may describe entities using different scales. To rectify this, one usually applies feature normalisation aiming at rescaling features so that none of them overpowers the others in the objective function of the selected clustering algorithm. In this paper, we argue that the rescaling procedure should not treat all features identically. Instead, it should favour the features that are more meaningful for clustering. With this in mind, we introduce a feature rescaling method that takes into account the within-cluster degree of relevance of each feature. Our comprehensive simulation study, carried out on real and synthetic data, with and without noise features, clearly demonstrates that clustering methods that use the proposed data normalization strategy clearly outperform those that use traditional data normalization.
\end{abstract} 
%

%
%
\section{Introduction}

The main goal of any clustering algorithm is to produce a set of clusters so that the entities belonging to a given cluster are homogeneous according to some criteria. Such algorithms have found use in a number of different fields such as bioinformatics, data mining, computer vision, etc. \cite{suzuki2006pvclust,panda2017diversity,berkhin2006survey,de2008clustering,deAmorimMak2016,wang2020novel,liu2020privacy,CordeirodeAmorimSMM2017}.

Given a data set $X=\{x_1, x_2, ..., x_n\}$ in which $x_i \in \mathbb{R}^m$, a partitional crisp clustering algorithm aims to produce a clustering $S=\{S_1, S_2, ..., S_k\}$, such that $\left|\bigcup_{l=1}^k S_l\right| = n$ and $S_i \cap S_j=\emptyset$ for $i,j = 1, 2, ..., k$ and $i \neq j$. It is common for partitional clustering algorithms to consider a distance between entities, such as for example the squared Euclidean distance given by $d(x_i, x_j) = \sum_{v=1}^m (x_{iv} - x_{jv})^2$. There are indeed different clustering approaches, including crisp and fuzzy data partitioning as well as hierarchical clustering \cite{aggarwal2014data,xu2015comprehensive}. Algorithms under the fuzzy clustering approach allow a given entity $x_i$ to be assigned to more than one cluster, each assignment has a degree of membership and these usually add to one. Hierarchical clustering algorithms seek to produce a hierarchy of clusters, usually presented as a tree. Thus, an entity $x_i$ may belong to more than one cluster when such clusters are at different levels of the tree. The approach presented in this paper relates to the crisp partitional clustering.  

A special attention in clustering should be given to data preprocessing. During this stage, one of the main concerns is to apply the most appropriate data normalisation technique. The $m$ features used to describe each entity $x_i \in X$ may be defined using different scales. Thus, a feature $v$ with a wider scale than any other feature in $X$ will have a higher contribution to clustering than any other individual feature. This may lead to poor cluster recovery, particularly if $v$ is not as meaningful as the other features in $X$. With this in mind, data sets are usually normalised so that no feature overpowers others. The main data normalisation techniques used in data mining are the range normalisation, min-max normalisation, rescaling to unit length, $z$-scores and robust $z$-scores.

Having a balanced data set with features defined at the same scale is certainly a good starting point. However, in this paper we maintain that data normalisation should not aim to treat all features identically. It should instead aim to favour features that are more meaningful for clustering. This is the main contribution of our study. We show that the cluster-specific feature weights produced by a particular algorithm can be interpreted as feature rescaling factors. We also show that rescaling a data set using these features weights leads to a better cluster recovery. Our approach is quite unusual because our feature weights are cluster-specific. Thus, in our method each feature $v$ is rescaled with $k$ factors, where $k$ is the number of clusters. 

The remainder of the paper is organised as follows: Section \ref{Sec:RelatedWork} describes relevant related work in both clustering and feature rescaling; Section \ref{Sec:NewAlg} presents our method in details, explaining in particular why it should work; Section \ref{Sec:Exp_results} describes our experimental set up and the results we obtained; Section \ref{Sec:Conclusion} finalises our paper by presenting the main contributions of this study.

\section{Related work}
\label{Sec:RelatedWork}

We begin this section by briefly reviewing relevant clustering algorithms. We then discuss popular methods for feature rescaling.

\subsection{Clustering algorithms}
$K$-means \cite{macqueen1967some} is arguably the most popular partitional clustering algorithm \cite{jain2010data,steinley2006k}. It aims to partition a data set $X$, containing $n$ entities, into $k$ non-overlapping clusters $S=\{S_1, S_2, ..., S_k\}$, so that $|\bigcup_{l=1}^k S_l| =n$. It does so by minimising the within-cluster sum of squares:

\begin{equation}
\label{Eq:kmeans}
	P(U,Z) = \sum_{l=1}^k \sum_{i=1}^n \sum_{v=1}^m u_{il} (x_{iv} - z_{lv})^2,
\end{equation}
where $u$ is a $n\times k$ binary matrix in which the value of $u_{il}$ indicates whether or not $x_i \in S_l$, and $z_l$ is the centroid of cluster $S_l$ (i.e., its centre of gravity). If the square Euclidean distance is used in (\ref{Eq:kmeans}), it becomes straightforward that 
$z_{lv} = |S_l|^{-1} \sum_{i=1}^n u_{il} x_{iv}$. We can summarise $k$-means in three steps:

\begin{enumerate}
	\item Select $k$ entities from $X$ uniformly at random and copy their values into the initial centroids $Z=\{z_1, z_2, ..., z_k\}$.
	\item Assign each $x_i \in X$ to the cluster $S_l$ whose centroid $z_l$ is the nearest to $x_i$ and update $u_{il}$ accordingly. If this step produces no change in $u$, then stop.
	\item Update $z_l$ to the component-wise mean of $x_i \in S_l$, for $l=1, 2, ..., k$.
\end{enumerate}
The algorithm above is guaranteed to converge. This is true for two reasons: (i) the number of possible partitions may be large, but it is finite and (ii) the output of (\ref{Eq:kmeans}) is monotonically decreasing, and by consequence no clustering is repeated. However, there is no guarantee the final clustering will be optimal. Due to its greedy nature, the final clustering found by $k$-means is usually only a local minima solution. Moreover, this clustering is highly dependent on the initial centroids (usually set up at random). In fact, finding the optimal clustering minimising (\ref{Eq:kmeans}) is an NP-hard problem, even for $k=2$ \cite{aloise2009np}.

In its original form, $k$-means has two parameters: the number of clusters ($k$), and the data set to be clustered ($X$). Finding the number of clusters in a data set is a non-trivial task that will clearly affect the final clustering $S$. The difficulty of identifying $k$ stems from the fact that from a mathematical perspective \textit{cluster} is a poorly defined term. In this paper we experiment solely on data sets for which $k$ is known, hence, avoiding this issue entirely. We direct interested readers to the recent related literature such as \cite{hennig2015true,unlu2019estimating,lord2017using,li2020cluster} and references therein.

There has been a considerable research effort aiming at designing algorithms capable of producing good initial centroids for $k$-means. Here, we discuss two methods that we find to be particularly relevant. We direct readers interested in a wider view to the following papers  \cite{yuan2004new,hatamlou2012search,erisoglu2011new,sun2002iterative,steinley2007initializing}, and references therein.

$K$-means++ \cite{arthur2007k} is a very popular implementation of $k$-means, which has become the default $k$-means program in MATLAB. This algorithm selects the first centroid at random from the entities, and the others using a weighted probability related to the distances between entities and their closest centroid already chosen. 

\begin{enumerate}
	\item Set $l=1$. Select an entity from $X$ uniformly at random and copy its values to $z_l$.
	\item Increment $l$ by one. Select an entity $x_j$ from $X$ at random, with probability $\frac{D(x_j)^2}{\sum_{i=1}^n D(x_i)^2}$ and copy its values to $z_l$.
	\item Repeat the steps above until $l=k$.
	\item Run $k$-means using the $\{z_1, z_2, ..., z_k\}$ as initial centroids.
\end{enumerate}
In the above, $D(x_i)$ represents the distance between $x_i$ and its nearest centroid. Experiments show that $k$-means++ has a faster convergence to a lower criterion output (\ref{Eq:kmeans}) than the traditional $k$-means algorithm \cite{arthur2007k}.

Intelligent $k$-means ($ik$-means) \cite{mirkin2012clustering} is another popular algorithm designed to determine good initial centroids for $k$-means. It does so by using the concept of anomalous patterns. We describe the main steps of this algorithm below.
\begin{enumerate}
	\item Find the entity $x_i \in X$ that is the farthest one from the data centre ($z_c$), and copy its values to $z_t$. 
	\item Run $k$-means on $X$ with two initial centroids, $z_t$ and $z_c$, leading to the clusters $S_t$ and $S_c$. During this clustering, do not allow $z_c$ to move at the centroid update step.
	\item If $|S_t|> \theta$ add $z_t$ to $Z^{\prime}$ and remove each $x_i \in S_t$ from $X$. If $|X|>0$, then go to Step 1.
	\item Run $k$-means on the whole original data set using the centroids in $Z^{\prime}$ as initial centroids.
\end{enumerate}
The above identifies a centroid $z_t$ and the related cluster $S_t$ by iteratively minimising:
\begin{equation}
\label{Eq:P(U,Z)_1}
P(U,Z) = \sum_{i=1}^n \sum_{v=1}^m u_{it}(x_{iv} - z_{tv})^2 +  \sum_{i=1}^n \sum_{v=1}^m u_{ic}(x_{iv} - z_{cv})^2.
\end{equation}

Given that clustering is usually done after data normalisation leading to the data centre ($z_c$) of zero, we can rewrite (\ref{Eq:P(U,Z)_1}) as follows:
\begin{equation}
\label{Eq:P(U,Z)_2}
P(U,Z) = \sum_{i=1}^n \sum_{v=1}^m u_{it} (x_{iv} - z_{tv})^2 +  \sum_{i=1}^n \sum_{v=1}^m u_{ic} x_{iv}^2.
\end{equation}

This anomalous pattern method identifies suitable initial centroids for $k$-means as well as the number of clusters $k$, and it does so quite successfully \cite{chiang2010intelligent}. In this paper we are not interested in finding the number of clusters in a data set, so when using this initialisation we set $\theta = 0$ and select the $k$ centroids in $Z^{\prime}$ with the largest cardinality.

The final clustering generated by $k$-means depends heavily on the initial centroids. Both $k$-means++ and $ik$-means attempt to identify good initial centroids, but unfortunately this is not the only weakness in $k$-means. The $k$-means criterion (\ref{Eq:kmeans}) assumes that every feature in the data set is equally relevant, which is hardly the case in real-life scenarios. With this in mind \cite{de2012minkowski} introduced the intelligent Minkowski weighted $k$-means ($imwk$-means) - an algorithm capable of successfully calculating within-cluster feature weights that improve cluster recovery \cite{de2016survey,melvin2016uncovering}. The Minkowski distance between entity $x_i \in S_l$ and centroid $z_l$ is defined by:
\begin{equation}
\label{Eq:MinkDist}
d(x_i, z_l) = \sum_v^m w_{lv}^p |x_{iv}-z_{lv}|^p,
\end{equation}
where $w_{lv}$ is the weight of feature $v$ at cluster $S_l$, and $p$ is a user-defined Minkowski exponent. The Minkowski distance is a generalisation of the Manhattan ($p = 1$), Euclidean ($p=2$) and Chebyshev ($p\rightarrow \infty$) distances. Hence, the value of $p$ can be used to adjust the distance bias of $imwk$-means. In two dimensions this bias is towards the cluster shapes of a diamond ($p=1$), circle ($p=2$), and square ($p\rightarrow \infty$). This is certainly an improvement over $k$-means given the latter uses the Euclidean distance, and is by consequence biased solely towards circular clusters. The $imwk$-means enjoys even further freedom in terms of cluster shape bias thanks to its use of cluster-specific feature weights. However, this is not to say $imwk$-means would be able to identify clusters of any shape. This, we believe, is an issue for any $k$-means based algorithm. Equation (\ref{Eq:MinkDist}) is in fact the $p^{th}$ power of the Minkowski distance, which is analogous to the use of the squared Euclidean distance in $k$-means. The distance (\ref{Eq:MinkDist}) leads to the new optimization criterion:
%
%
\begin{equation}
	P(U,Z, W) = \sum_{l=1}^k \sum_{i=1}^n \sum_{v=1}^m u_{il} w_{lv}^p |x_{iv} - z_{lv}|^p.
\label{Eq:imwk}
\end{equation}

The minimisation of (\ref{Eq:imwk}) subject to a crisp clustering and $\sum_{v=1}^m w_{lv}=1$ for $l=1, 2, ..., k$ implies:
\begin{equation}
\label{Eq:MinkWeights}
w_{lv} = \frac{1}{\sum_{j=1}^m \left[\frac{D_{lv}}{D_{lj}} \right]^\frac{1}{p-1}},
\end{equation}
where $D_{lv} = \sum_{i=1}^n  u_{il} |x_{iv} - z_{lv}|^p$. We can minimise (\ref{Eq:imwk}) by adding an extra step to $k$-means. We refer to this as the Minkowski weighted $k$-means ($mwk$-means). The steps of this algorithm are as follows:
\begin{enumerate}	
	\item Select $k$ entities from $X$ uniformly at random, copy their values to the initial centroids $Z=\{z_1, z_2, ..., z_k\}$. Set each $w_{lv} = m^{-1}$.
	\item Assign each $x_i \in X$ to the cluster $S_l$ whose centroid $z_l$ is the nearest to $x_i$ as per (\ref{Eq:MinkDist}), and update $u_{il}$ accordingly. If this step produces no change in $u$, stop.
	\item Update each $z_l$ to the Minkowski centre of its cluster $S_l$ (see below).
	\item Update each $w_{lv}$ as per (\ref{Eq:MinkWeights}). Go back to Step 2.
\end{enumerate}

The Minkowski centre for feature $v$ at cluster $S_l$ is the value $\mu$ that minimises $\gamma_v(\mu) = \sum_{i=1}^n u_{il} |x_{iv} - \mu|^p$. Notice that at $p\geq 1$, $\gamma(\mu)$ is a U-shaped curve with a minimum in the interval $\left[ min(x_v), max(x_v)\right]$. We can then minimise $\gamma(\mu)$ using standard methods for convex optimisation. For instance, to find the Minkowski centre for feature $v$ at cluster $S_l$ we can start with $\mu = |S_l|^{-1} \sum_{i=1}^n u_{il} x_{iv}$. We then move $\mu$ by a fixed amount ($0.001$, say) per step to the side that reduces $\gamma_v$. The $imwk$-means includes a Minkowski-based $ik$-means initialisation designed to find good initial centroids as well as good feature weights, as we can see below.
\begin{enumerate}
	\item Set $z_c$ to be the Minkowski centre of $X$, and each $w_{lv} = m^{-1}$.
	\item Find the entity $x_i \in X$ that is the farthest from $z_c$ using (\ref{Eq:MinkDist}) and copy its values to $z_t$. 
	\item Run $mwk$-means using $z_c$ and $z_t$ as initial centroids, leading to the clusters $S_c$ and $S_t$. In Step 3 of $mwk$-means do not allow $z_c$ to move.
	\item Add $z_t$ to $Z^{\prime}$ and $w$ to $W^{\prime}$. 
	\item Remove all entities $x_i \in S_t$ from $X$. If $|X|>0$ go to Step 2.
	\item Keep in $Z^{\prime}$ and $W^{\prime}$ only the elements related to the $k$ clusters with the highest cardinality.
	\item Run $mwk$-means on the original data set $X$ initialised with the centroids in $Z^{\prime}$ and weights in $W^{\prime}$.
\end{enumerate}

The $imwk$-means algorithm clearly supports the intuitive idea that a feature $v$ may be more meaningful to one cluster than to another. We model this using $w_{lv}$ to set the degree of relevance of feature $v$ at cluster $S_t$.
\subsection{Feature rescaling}
\label{SSec:FeatureRescaling}
Clustering algorithms usually require feature rescaling in the data preprocessing step. The general idea is to balance the values of features so that those with a higher scale do not overpower others. For instance, when clustering individuals the feature weight (measured in kilograms) will have a higher contribution to the criterion (\ref{Eq:kmeans}) than the feature height (measured in meters). This happens because in absolute values the weight of humans tends to be considerably larger than their height (in other words, the variance of weight is larger than that of height). Once all features present values on a common scale, the data set can be clustered.

There are different feature rescaling methods that can be applied during the data preprocessing step of clustering. Here, we focus on the most popular of them.\\
\\\textit{Z-scores}\\
The $z$-score normalisation is arguably the most popular approach of feature rescaling. The $z$-score of $x_{iv}$ is given by:

\begin{equation}
\label{Eq:zscore}
x'_{iv} = \frac{x_{iv} - \bar{x}_v}{\sigma_v},
\end{equation}
where $\bar{x}_v$ and $\sigma_v$ are the mean and standard deviation of feature $v$, respectively. The standardised $x_{iv}'$ represents the number of standard deviations by which the original $x_{iv}$ is above $\bar{x}_v$. A popular extension of this method is the robust $z$-score normalisation in which the mean is replaced by the median and the standard deviation by MAD (Median Absolute Deviation). This method is more robust than $z$-scores in the presence of outliers.\\\\
\textit{Range normalisation}\\
The range normalisation is a popular alternative to the $z$-score, particularly in cluster analysis . The normalised value of $x_{iv}$ is computed as follows:
\begin{equation}
\label{Eq:Stand}
x'_{iv} = \frac{x_{iv} - \bar{x}_v}{max(x_v) - min(x_v)},
\end{equation}
where $max(x_v)$ and $min(x_v)$ are the maximum and minimum values of feature $v$, respectively. There is a crucial difference between the range normalisation and $z$-scores, the latter is biased toward unimodal distributions. This is probably easier to explain with an example. Let us take two features, a unimodal $v_1$ and a multimodal $v_2$. The standard deviation of $v_2$ is likely to be higher than that of $v_1$. Thus, the $z$-score value of $v_1$ will be higher than that of $v_2$ even though $v_2$ has a better cluster information.\\\\
\textit{Min-max normalisation}\\
This is arguably the simplest method one can use to normalise features. It rescales the features of a given data set to the interval $[0, 1]$:
\begin{equation}
\label{Eq:Stand_minmax}
x'_{iv} = \frac{x_{iv} - min(x_v)}{max(x_v) - min(x_v)}.
\end{equation}
\textit{Rescaling to unit length}\\
A feature $v$ can also be interpreted as being a vector. Thus, it is possible to normalise the components of $v$ so that this vector has a length of one. In this case, the normalised value of $x_{iv}$ is obtained as follows:
\begin{equation}
\label{Eq:Stand_veclength}
x'_{iv} = \frac{x_{iv}}{||x||},
\end{equation}
where $||x_v|| = \sqrt{\sum_{i=1}^n x_{iv}^2}$ is the Euclidean length of feature $v$.

\section{Clustering with feature rescaling factors}
\label{Sec:NewAlg}
%
Feature rescaling methods such as those discussed in Section \ref{SSec:FeatureRescaling} are certainly a good starting point, but we maintain their application cannot be the final step of the data preprocessing stage. This stage should not aim at treating all features equally, but should instead favour features that have a higher degree of relevance for clustering. With this in mind, let us analyse the weights generated by $imwk$-means. Our objective here is to get a set of weights minimising (\ref{Eq:imwk}) subject to $\sum_{v=1}^m w_{lv}=1$ for $l=1, 2, ..., k$ within a crisp clustering criterion (i.e., $S_i \cap S_j$ for $i,j=1, 2, ..., k$ and $i \neq j$). Given that  $D_{lv} = \sum_{i=1}^n  u_{il} |x_{iv} - z_{lv}|^p$, we can rewrite the function to be minimised (\ref{Eq:imwk}) as follows:

\begin{equation*}
	P(U,Z, W) = \sum_{v=1}^m  \sum_{l=1}^k w_{lv}^p D_{lv}.
\end{equation*}

Since we calculate feature weights one cluster at a time and $\sum_{v=1}^m w_{lv}=1$ for $l=1, 2, ..., k$, the following Lagrangian function can be formulated:
\begin{equation*}
\mathcal{L}(W, \lambda)= \sum_{v=1}^m w_{lv}^p D_{lv} + \lambda\left(1-\sum_{v=1}^m w_{lv} \right).
\end{equation*}
The partial derivatives of $\mathcal{L}$ with respect to $w_{lv}$ and $\lambda$ can be equated to 0. They are as follows:
\begin{equation}
\label{Eq:FirstPartial}
\frac{\partial \mathcal{L}}{\partial w_{lv}} = pw_{lv}^{p-1}D_{lv} - \lambda = 0,
\end{equation}
\begin{equation}
\label{Eq:SecondPartial}
\frac{\partial \mathcal{L}}{\partial \lambda} = 1-\sum_{v=1}^m w_{lv} = 0,
\end{equation}
respectively. Equation (\ref{Eq:FirstPartial}) leads to:
\begin{equation}
\label{Eq:FirstPartial_2}
w_{lv} = \left(\frac{\lambda}{pD_{lv}}\right)^\frac{1}{p-1}.
\end{equation}
Substituting (\ref{Eq:FirstPartial_2}) into (\ref{Eq:SecondPartial}), we obtain:
\begin{equation*}
\sum_{v=1}^m \left( \frac{\lambda}{pD_{lv}}\right)^\frac{1}{p-1}=1,
\end{equation*}
leading to:
\begin{equation*}
\left(\lambda\right)^\frac{1}{p-1} = \frac{1}{\sum_{v=1}^m \left(\frac{1}{pD_{lv}}\right)^\frac{1}{p-1}}.
\end{equation*}
and by consequence to Equation (\ref{Eq:MinkWeights}). The above demonstrates that the features generated by $imwk$-means are in fact quite specific: they minimise (\ref{Eq:imwk}) and model the within-cluster degree of relevance of each feature. Thus, they can be used to rescale a data set in a rather unconventional way. Given $u$ and feature weights $w$, we can rescale a data set by setting:
\begin{equation}
\label{Eq:ReScale}
x'_{iv} = \sum_{l=1}^k u_{il} x_{iv} w_{lv}.
\end{equation}

This is unconventional because a given feature $v$ is rescaled with $k$ different factors $w_{1}, w_{2}, ..., w_{k}$. We propose to improve cluster recovery with feature rescaling factors following the steps below.

\begin{enumerate}
	\item Standardise the data set $X$ using either (\ref{Eq:zscore}),  (\ref{Eq:Stand}), (\ref{Eq:Stand_minmax}), or (\ref{Eq:Stand_veclength}).
	\item Find a clustering $U$ and weights $W$ by applying the $imwk$-means to the standardised data set.
	\item Rescale $X$ using $U$ and $W$ by applying (\ref{Eq:ReScale}).
	\item Apply $imwk$-means to the rescaled data set, leading to the clustering $U^{\prime}$.
\end{enumerate}
One should note that rescaling a feature with (\ref{Eq:ReScale}) only makes sense because the weights in $W$ minimise (\ref{Eq:imwk}). The $imwk$-means requires a user-defined parameter $p$ and its rescaled version requires two values of $p$. The value of $p_1$ is used to generate the clustering $U$ and weights $W$ (Step 2) and the value $p_2$ is used in clustering of the rescaled data set (Step 4). Of course, one could set $p_1 = p_2$.

\begin{figure}[t]
    \centering
    \begin{subfigure}[b]{0.4\textwidth}
        \includegraphics[width=\textwidth]{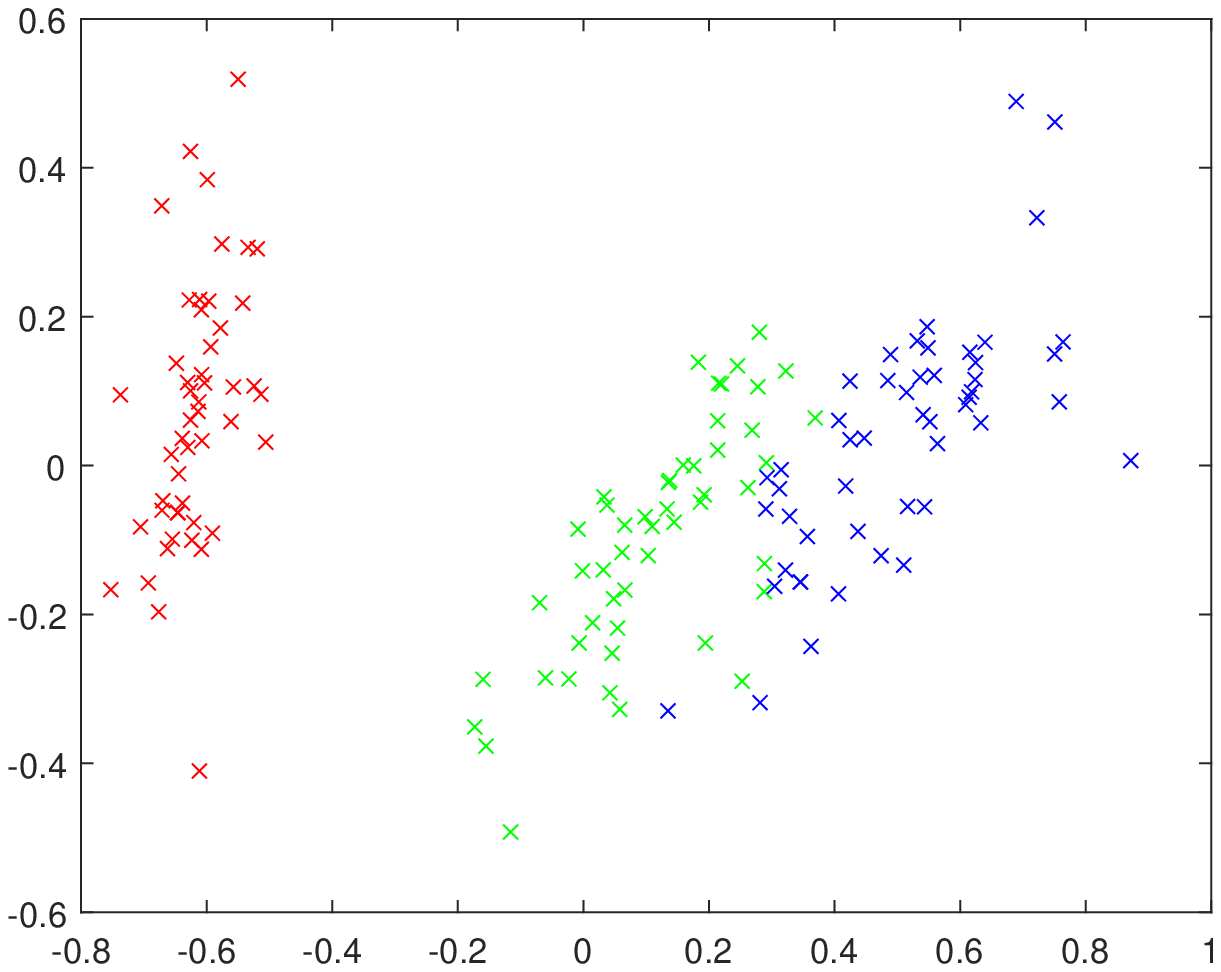}
        \caption{The Iris data set.}
        \label{Fig:Iris_Range}
    \end{subfigure}
    \begin{subfigure}[b]{0.4\textwidth}
        \includegraphics[width=\textwidth]{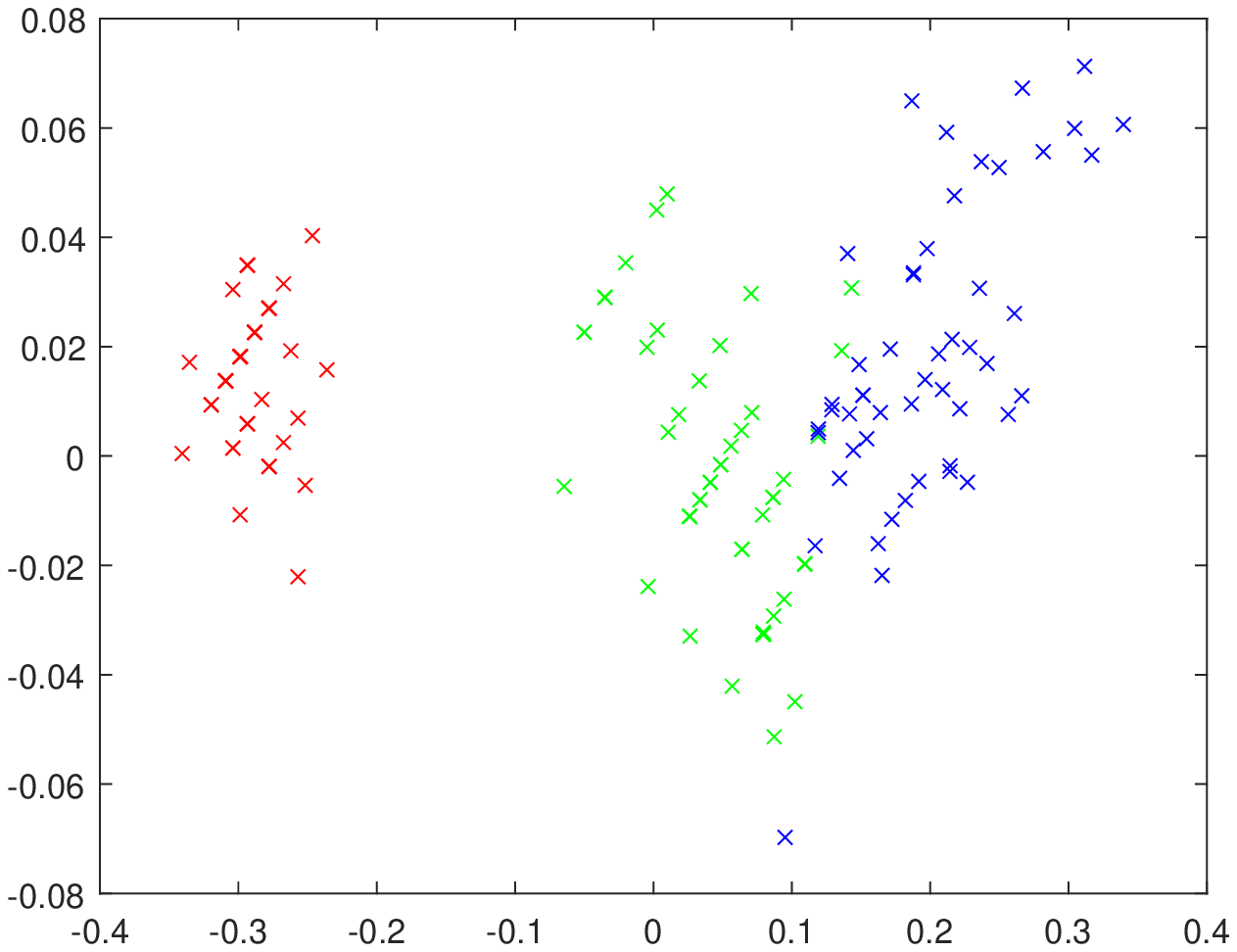}
        \caption{The rescaled Iris data set.}
        \label{Fig:Iris_Range_ReScaled}
    \end{subfigure}
    \begin{subfigure}[b]{0.4\textwidth}
        \includegraphics[width=\textwidth]{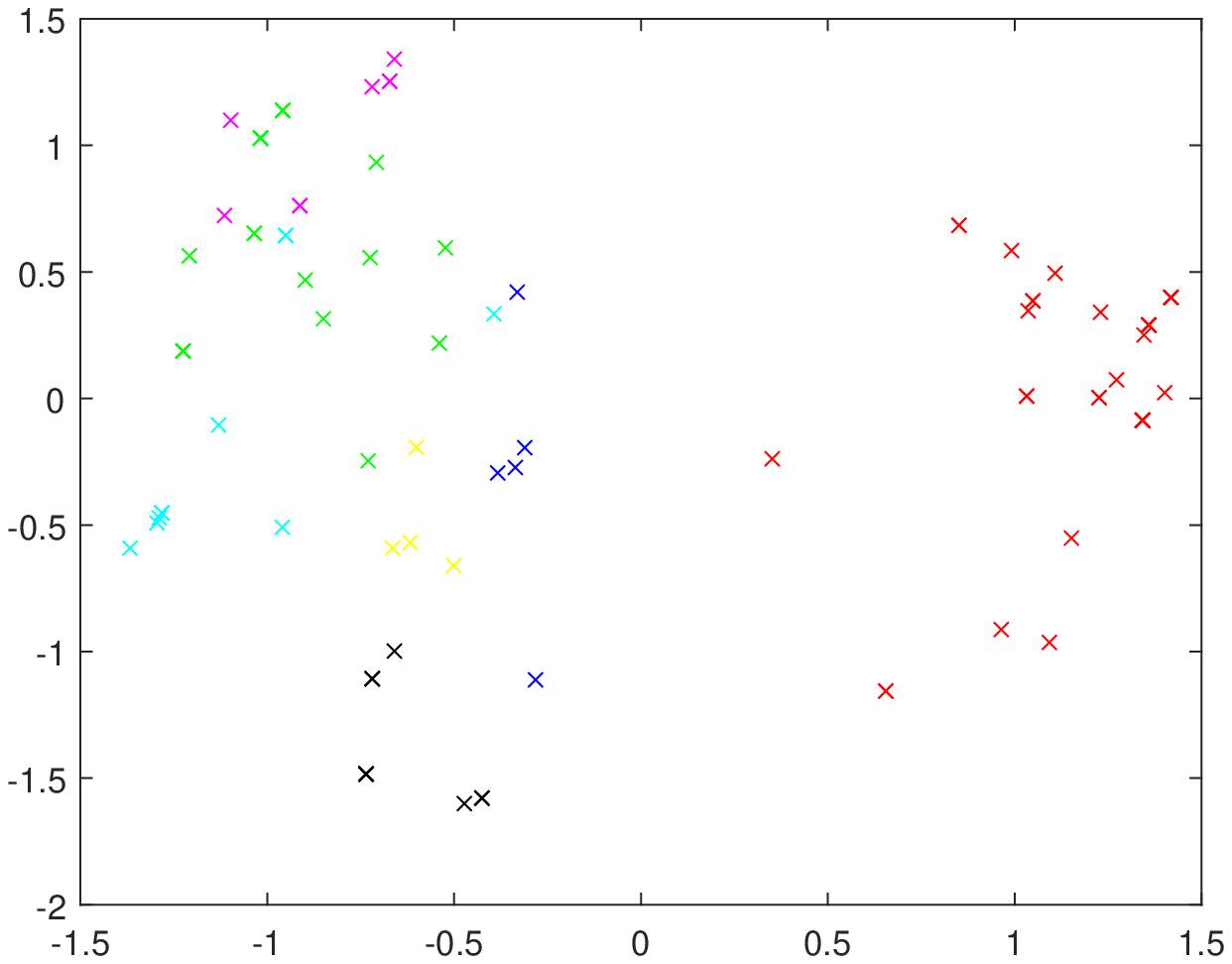}
        \caption{The Zoo data set.}
        \label{Fig:Zoo_Range}
    \end{subfigure}
    \begin{subfigure}[b]{0.4\textwidth}
        \includegraphics[width=\textwidth]{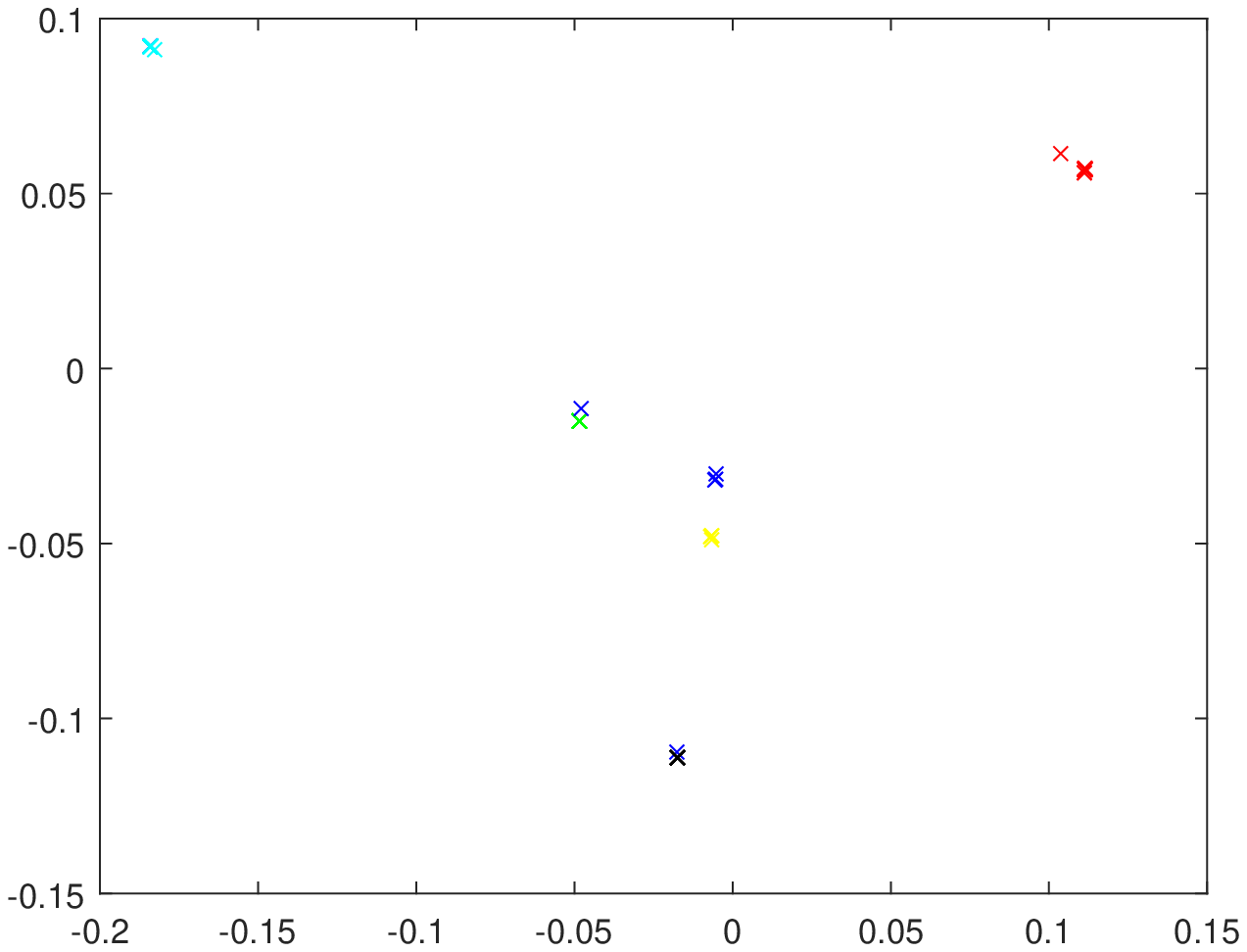}
        \caption{The rescaled Zoo data set.}
        \label{Fig:Zoo_Range_ReScaled}
    \end{subfigure}
    \caption{The Iris and Zoo data sets plotted over their first and second principal components. Figures \ref{Fig:Iris_Range} and \ref{Fig:Zoo_Range} show the data sets normalised using the range normalisation. Figures \ref{Fig:Iris_Range_ReScaled} and \ref{Fig:Zoo_Range_ReScaled} show these data sets rescaled using our method.}
     \label{Fig:rescaling_examples}
\end{figure}

Figure \ref{Fig:rescaling_examples} exemplifies the effect our rescaling method can have on data sets (for details on the data sets themselves, see Section \ref{SubSec:setup}). In this, it is possible to see that clusters tend to become more compact (see the read and blue clusters in the Iris data set, and all clusters in the Zoo data set), and more separable. Section \ref{Sec:Exp_results} takes this further by showing clear improvements in terms of cluster recovery.

According to Formula (\ref{Eq:ReScale}), the rescaling part of the algorithm has a linear time complexity with respect to the number of clusters $k$, the number of features $m$, and the number of entities $n$. Thus, our rescaling would not increase the asymptotic time complexity of the $k$-means or $imwk$-means algorithms.
\section{Experimental results}
\label{Sec:Exp_results}
The data preprocessing stage is crucial for any clustering algorithm. At this stage, features are usually put on the same scale so that none overpowers any of the others. The main objective of our experiments is to demonstrate that a rescaling favouring meaningful features leads to better cluster recovery. This happens because the feature rescaling factors generated by our method minimise the sum of within-cluster distances (see Section \ref{Sec:NewAlg}). Hence, we envisage that our method could be used at the data preprocessing step of any $k$-means-based clustering algorithm.

In this section, we experiment with four algorithms: (i) $k$-means++, giving the reader a clear idea of a generally accepted baseline for each data set; (ii) $imwk$-means, allowing us to generate the feature rescaling factors; (iii) $k$-means++ with a data set rescaled using our method; (iv) $imwk$-means with a data set rescaled using our method. We show that our rescaling method allows one to improve cluster recovery. 

We divided the remainder of this section into two parts. First, we explain the details related to the data sets used in our experiments. Then, we present and discuss the obtained results for both synthetic and real data.
\subsection{Experimental setup}
\label{SubSec:setup}
We experimented with a total of 600 synthetic data sets generated under 12 different configurations (see Table \ref{Tab:Datasets}), with and without added noise, and 18 real-world data sets (see Table \ref{Tab:real_world_datasets}) downloaded from the popular UCI Machine learning repository \cite{Dua:2017}. 

The synthetic data sets contain spherical Gaussian clusters. Their covariance matrices are diagonal, with the same diagonal value $\sigma^2$, generated at each cluster randomly between $0.5$ and $1.5$. Each centroid component was generated independently from a Gaussian distribution with zero mean and unity variance. Each cluster has a cardinality taken from a uniform distribution, subject to a minimum of 20 entities. We initially generated 50 data sets under each of the following configurations: (i) 1000 entities over six features partitioned into three clusters (1000x6-3); (ii) 1000 entities over 12 features partitioned into six clusters (1000x12-6); (iii) 1000 entities over 20 features partitioned into ten clusters (1000x20-10).

For each synthetic data set containing $m$ features, we have generated two data sets including $\lceil \frac{m}{2}\rceil$ noise features (leading to a total of $\lceil 1.5m\rceil$ features) and one other data set with within cluster noise. Here, we experiment with three models of noise. In the first, we considered a noise feature (NF) as a feature containing solely uniformly random values. In the second, we considered a noise feature as one containing random values from a Gaussian distribution (NNF). In our third noise model we selected 50\% of the $m \times k$ feature segments uniformly at random, and then substituted the selected segments with uniformly random values - creating within cluster noise (WCN). This approach has quadrupled the number of data sets tested in our experiments. 

Regarding our real-world data sets, we replaced any categorical feature $v$ containing $t$ categories with $t$ binary features. For any given entity only one of the new $t$ features was set to one, that which represented the original value of $v$. In all of our experiments we have rescaled the features in a data set using one of the methods described in Section \ref{SSec:FeatureRescaling}. 

Given that we knew the true labels for all our data sets (i.e. their true structure was known), we were able to assess the quality of cluster recovery using an external validation index. In the case of clustering, one of the strongest contenders is the Adjusted Rand Index (ARI) \cite{rand1971objective}. The ARI between the clusterings $S=\{S_1, S_2, ..., S_k\}$ and $U=\{U_1, U_2, ..., U_r\}$ is defined as follows: 
\begin{equation}
ARI(S, U) = \frac{\sum_{ij} \binom{n_{ij}}{2} - [\sum_i \binom{a_i}{2} \sum_j \binom{b_j}{2}] / \binom{n}{2} }{ \frac{1}{2} [\sum_i \binom{a_i}{2} + \sum_j \binom{b_j}{2}] - [\sum_i \binom{a_i}{2} \sum_j \binom{b_j}{2}] / \binom{n}{2} },
\end{equation}
where $n_{ij} = |S_i \cap U_j|$, $a_i = \sum_{j=1}^r |S_i \cap U_j|$, $b_j = \sum_{i=1}^k |S_i \cap U_j|$. 
%
%
%
\begin{table}\footnotesize
\begin{center}
\caption{50 data sets for each of the configurations below were considered. Each data set contained Gaussian clusters with different spreads and cardinalities. We added noise features containing uniformly random values (NF), normal random values (NNF), or within cluster noise (WCN), to the data sets in some configurations.}
\label{Tab:Datasets}
\tabcolsep=0.08cm
\begin{tabular}{lcccccccc}
&Noise&&\multicolumn{4}{c}{Features}&&Clusters\\
\cline{4-7}
&(\%)&&normal&uniform&original&total&&($k$)\\
&&&noise&noise&&($m$)\\
1000x6-3&0.00&&0&0&6&6&&3\\
1000x12-6&0.00&&0&0&12&12&&6\\
1000x20-10&0.00&&0&0&20&20&&10\\
\cline{1-9}
1000x6-3 +3NF&33.33&&0&3&6&9&&3\\
1000x12-6 +6NF&33.33&&0&6&12&18&&6\\
1000x20-10 +10NF&33.33&&0&10&20&30&&10\\
\cline{1-9}
1000x6-3 +3NNF&33.33&&3&0&6&9&&3\\
1000x12-6 +6NNF&33.33&&6&0&12&18&&6\\
1000x20-10 +10NNF&33.33&&10&0&20&30&&10\\
\cline{1-9}
1000x6-3 WCN&50.00&&0&0&6&6&&3\\ 
1000x12-6 WCN&50.00&&0&0&12&12&&6\\ 
1000x20-10 WCN&50.00&&0&0&20&20&&10\\
\end{tabular}
\end{center}
\end{table}
\begin{table}[t]\footnotesize
\begin{center}
\caption{Details of all real-world data sets we experiment with. The number of features shown below is that obtained after handling categorical features (see Section \ref{SubSec:setup}).}
\label{Tab:real_world_datasets}
\tabcolsep=0.07cm
\begin{tabular}{lccc}
&Entities&Features&Clusters\\
&($n$)&($m$)&($k$)\\
Australian CC&690&42&2\\ 
Balance&625&4&3\\ 
Breast cancer&699&9&2\\ 
Car evaluation&1,728&21&4\\ 
Mines vs rocks&208&28&2\\ 
Ecoli&336&7&8\\ 
Glass&214&9&6\\ 
Heart&270&25&2\\ 
Ionosphere&351&33&2\\ 
Iris&150&4&3\\ 
Lung cancer&32&56&3\\ 
Musk&476&166&2\\ 
Parkinsons&195&22&2\\ 
Soya&47&58&4\\ 
Teaching assistant&151&56&3\\ 
Tic-tac-toe&958&27&2\\ 
Wine&178&13&3\\ 
Zoo&101&16&7\\ 
\end{tabular}
\end{center}
\end{table}

\subsection{Results and analysis}

In our first set of experiments, we compared the four feature rescalings presented in Section \ref{SSec:FeatureRescaling} using the data sets described in Section \ref{SubSec:setup}. Table \ref{Tab:Results_kmpp} shows the average ARI values between the clusterings generated by $k$-means++ on Synthetic data sets using these four feature rescaling approaches. Table \ref{Tab:Results_kmpp_rw} reports the results of similar experiments made on the real-world data sets. Given that $k$-means usually returns different clustering solutions for different random initial partitions, we ran $k$-means++ 100 times per data set (notice that we tested 50 different data sets per parameter configuration in the case of synthetic data sets). 

Some interesting patterns could be found when observing the results reported in Table \ref{Tab:Results_kmpp}. For example, the range normalisation produced slightly better results than the other approaches for the original data sets (no noise features) as well as for the data sets containing normal noise values (NNF). We certainly acknowledge the differences in ARI are very small, but they still put the range normalisation as at least one of the best in these scenarios. However, for the data sets containing uniformly random noise features (NF) or within cluster noise (WCN), it is the $z$-score normalisation that produced the best results, while the range-based normalisations (min-max and range normalisation) performed poorly. The main reason for this is that the features containing uniformly random values do not correspond to any cluster structure. Thus, the standard deviation of such noise features is likely to be higher than that of the original features. Given that the $z$-score (\ref{Eq:zscore}) formula includes a division by the standard deviation, uniformly distributed random noise features have lower $z$-score values. By consequence, these noise features have a lower contribution to the clustering (\ref{Eq:kmeans}). 

The above does not imply that $z$-scores is the best normalisation method overall. Table \ref{Tab:Results_kmpp_rw} shows that $z$-scores lead to the highest ARI only for 4 of our 18 real-world data sets. For the real-world data, it is the range normalisation that provided the best performance, followed by min-max. 
\begin{table}[t]\footnotesize
\begin{center}
\caption{Average ARI and standard deviation values for the clusterings found by $k$-means++ on synthetic data sets. We ran $k$-means++ 100 times per data set. There were 50 data sets per parameter configuration. Each of the four main columns presents the results for a different normalisation approach.}
\label{Tab:Results_kmpp}
\tabcolsep=0.07cm
\begin{tabular}{lcccccccccccc}
&\multicolumn{2}{c}{min-max}&&\multicolumn{2}{c}{range norm}&&\multicolumn{2}{c}{$z$-score}&&\multicolumn{2}{c}{unit length}\\
\cline{2-3}
\cline{5-6}
\cline{8-9}
\cline{11-12}
&ARI&std&&ARI&std&&ARI&std&&ARI&std\\
1000x6-3&0.5145&0.22&&0.5198&0.22&&0.5060&0.21&&0.4821&0.23\\
1000x12-6&0.6338&0.18&&0.6356&0.18&&0.6336&0.17&&0.6222&0.18\\
1000x20-10&0.7680&0.12&&0.7703&0.12&&0.7708&0.11&&0.7712&0.12\\ 
\cline{1-12}
1000x6-3 +3NF&0.0365&0.11&&0.0371&0.11&&0.4550&0.21&&0.4014&0.22\\ 
1000x12-6 +6NF&0.0994&0.11&&0.0997&0.11&&0.5820&0.17&&0.5513&0.18\\ 
1000x20-10 +10NF&0.1706&0.10&&0.1708&0.10&&0.7233&0.14&&0.7184&0.12\\  
\cline{1-12}
1000x6-3 +3NNF&0.4735&0.23&&0.4748&0.23&&0.4690&0.21&&0.4102&0.22\\
1000x12-6 +6NNF&0.5837&0.17&&0.5852&0.17&&0.5840&0.17&&0.5552&0.18\\ 
1000x20-10 +10NNF&0.7277&0.13&&0.7286&0.13&&0.7278&0.12&&0.7205&0.12\\ 
\cline{1-12}
1000x6-3 WCN&0.0594&0.06&&0.0597&0.06&&0.1645&0.09&&0.1473&0.09\\ 
1000x12-6 WCN&0.0926&0.06&&0.0920&0.06&&0.1835&0.07&&0.1767&0.07\\ 
1000x20-10 WCN&0.1051&0.03&&0.1049&0.03&&0.1735&0.05&&0.1698&0.05\\
\end{tabular}
\end{center}
\end{table}
\begin{table}[t]\footnotesize
\begin{center}
\caption{Average ARI and standard deviation values for the clusterings found by $k$-means++ on real-world data sets. We ran $k$-means++ 100 times per data set. Each of the four main columns presents the results for a different normalisation approach.}
\label{Tab:Results_kmpp_rw}
\tabcolsep=0.07cm
\begin{tabular}{lcccccccccccc}
&\multicolumn{2}{c}{min-max}&&\multicolumn{2}{c}{range norm}&&\multicolumn{2}{c}{$z$-score}&&\multicolumn{2}{c}{unit length}\\
\cline{2-3}
\cline{5-6}
\cline{8-9}
\cline{11-12}
&ARI&std&&ARI&std&&ARI&std&&ARI&std\\
\cline{1-12}
Australian CC&0.297&0.20&&0.308&0.20&&0.261&0.23&&0.169&0.22\\ 
Balance&0.141&0.04&&0.132&0.04&&0.130&0.04&&0.136&0.05\\ 
Breast cancer&0.839&0.00&&0.839&0.00&&0.826&0.01&&0.806&0.01\\ 
Car evaluation&0.057&0.05&&0.062&0.05&&0.020&0.03&&0.009&0.01\\ 
Mines vs rocks&-0.004&0.00&&-0.004&0.00&&0.001&0.00&&0.003&0.00\\ 
Ecoli&0.452&0.07&&0.457&0.07&&0.513&0.09&&0.446&0.08\\ 
Glass&0.190&0.04&&0.185&0.04&&0.180&0.06&&0.137&0.04\\ 
Heart&0.335&0.12&&0.341&0.10&&0.321&0.16&&0.260&0.20\\ 
Ionosphere&0.164&0.04&&0.168&0.03&&0.162&0.03&&0.107&0.04\\ 
Iris&0.669&0.10&&0.680&0.09&&0.591&0.07&&0.872&0.08\\ 
Lung cancer&0.119&0.07&&0.115&0.07&&0.067&0.07&&0.146&0.09\\ 
Musk&0.003&0.00&&0.003&0.00&&0.003&0.00&&0.001&0.00\\ 
Parkinsons&0.049&0.00&&0.049&0.00&&-0.088&0.01&&-0.071&0.00\\ 
Soya&0.779&0.20&&0.769&0.20&&0.823&0.19&&0.752&0.20\\ 
Teaching assistant&0.030&0.01&&0.031&0.01&&0.008&0.01&&0.004&0.01\\ 
Tic-tac-toe&0.016&0.02&&0.016&0.02&&0.014&0.01&&0.012&0.01\\ 
Wine&0.844&0.05&&0.838&0.07&&0.863&0.10&&0.761&0.10\\ 
Zoo&0.710&0.12&&0.684&0.13&&0.707&0.10&&0.660&0.11\\ 
\end{tabular}
\end{center}
\end{table}

Given the results in Tables \ref{Tab:Results_kmpp} and \ref{Tab:Results_kmpp_rw}, we decided to study in more details the range and $z$-score normalisations. We then devised a set of experiments aiming to find out whether there are parameters for the rescaled $imwk$-means and rescaled $k$-means++ that would lead to better cluster recovery than the best possible clustering by $imwk$-means and the expected $k$-means++ clustering, respectively. As the $k$-means++ algorithm is among the most popular variations of $k$-means \cite{arthur2007k}, it would be important to propose a method that outperforms it, regardless of the data normalization being used. 

Table \ref{Tab:Results_BestPossible_std} presents the results of these experiments on our synthetic data sets, all normalised using the range normalisation. There are 50 data sets for each configuration, this is the reason why we present the standard deviations for all the four competing algorithms. In this table we can observe a clear pattern. The average ARI given by the rescaled $imwk$-means and rescaled $k$-means++ in this experiment are higher than that of $imwk$-means and $k$-means++, respectively. In the case of the rescaled $imwk$-means this pattern becomes even clearer as the number of features and clusters increases. It is also interesting to see that the standard deviations of the ARIs obtained for the rescaled $imwk$-means is slightly lower than those of $imwk$-means and $k$-means++ in the majority of cases. Unsurprisingly the cluster recovery improvements provided by $imwk$-means and its rescaled version are higher for data sets containing noise features. This is a fair expectation for feature weighting algorithms. It is interesting to see that for $imwk$-means and its rescaled version we also have the following trend: the higher the ARI, the lower the standard deviation is. 

Table \ref{Tab:Results_BestPossible_zstd} reports the results for the experiments with synthetic data sets, in which the data were normalised using $z$-scores. The general patterns are still the same. The rescaled $imwk$-means and rescaled $k$-means++ provided better results than $imwk$-means and $k$-means++, respectively. The only major difference is that now $k$-means++ produced better results for the data sets containing noise features composed of uniformly random values (NF). We have explained the reason for this in the beginning of this section. The $k$-means++ results for the data sets containing within cluster noise (WCN) were also somewhat better, but still poor overall.

The superiority of our method is further confirmed by the results in Table \ref{Tab:Results_BestPossible_std_rw}. The results for the 18 real-world data sets suggest that our method is able to reach an equal or higher value of ARI in the vast majority of cases.

\begin{table}[t]\footnotesize
\begin{center}
\caption{A comparison between the ARI values for $imwk$-means and its rescaled version, supplied with good values of the exponents $p_1$ and $p_2$, and the expected ARI given by $k$-means++. The presented results are the averages over 50 data sets for each of the configurations below, each normalised using the range normalisation.}
\label{Tab:Results_BestPossible_std}
\tabcolsep=0.08cm
\begin{tabular}{lccccccccccc}
&&&&\multicolumn{2}{c}{Rescaled}&&&&&\multicolumn{2}{c}{Rescaled}\\
&\multicolumn{2}{c}{\textit{k}-means++}&&\multicolumn{2}{c}{\textit{k}-means++}&&\multicolumn{2}{c}{\textit{imwk}-means}&&\multicolumn{2}{c}{\textit{imwk}-means}\\
\cline{2-3}
\cline{5-6}
\cline{8-9}
\cline{11-12}
&ARI&std&&ARI&std&&ARI&std&&ARI&std\\
1000x6-3&0.5198&0.224&&0.5643&0.218&&0.5794&0.223&&0.6474&0.191\\ 
1000x12-6&0.6356&0.177&&0.6713&0.155&&0.7376&0.174&&0.7958&0.136\\ 
1000x20-10&0.7703&0.121&&0.7938&0.085&&0.9070&0.076&&0.9390&0.055\\ 
\cline{1-12}
1000x6-3 +3NF&0.0371&0.112&&0.4984&0.258&&0.5541&0.283&&0.6781&0.198\\ 
1000x12-6 +6NF&0.0997&0.111&&0.6330&0.166&&0.7543&0.156&&0.8307&0.116\\ 
1000x20-10 +10NF&0.1708&0.099&&0.7697&0.086&&0.8239&0.082&&0.9356&0.041\\ 
\cline{1-12}
1000x6-3 +3NNF&0.4748&0.225&&0.5512&0.211&&0.5864&0.215&&0.6495&0.200\\ 
1000x12-6 +6NNF&0.5852&0.172&&0.6520&0.153&&0.7358&0.174&&0.7832&0.156\\ 
1000x20-10 +10NNF&0.7286&0.125&&0.7777&0.087&&0.9050&0.059&&0.9381&0.041\\ 
\cline{1-12}
1000x6-3 WCN&0.0597&0.059&&0.2471&0.154&&0.3916&0.194&&0.4597&0.162\\ 
1000x12-6 WCN&0.0920&0.062&&0.3944&0.155&&0.6669&0.157&&0.6811&0.139\\ 
1000x20-10 WCN&0.1049&0.032&&0.5623&0.107&&0.8481&0.049&&0.8868&0.048\\ 

\end{tabular}
\end{center}
\end{table}
\begin{table}[t]\scriptsize
\begin{center}
\caption{A comparison on real-world data sets between the ARI values for $imwk$-means and its rescaled version, supplied with good values of the exponents $p_1$ and $p_2$, and the expected ARI given by $k$-means++.}
\label{Tab:Results_BestPossible_std_rw}
\tabcolsep=0.02cm
\begin{tabular}{lccccccccccccccccccc}
&\multicolumn{9}{c}{range norm}&&\multicolumn{9}{c}{$z$-score}\\
\cline{2-10}
\cline{12-20}
&&&&\multicolumn{2}{c}{Rescaled}&&&&Rescaled&&&&&\multicolumn{2}{c}{Rescaled}&&&&Rescaled\\\
&\multicolumn{2}{c}{$k$-means++}&&\multicolumn{2}{c}{$k$-means++}&&$imwk$&&$imwk$&&\multicolumn{2}{c}{$k$-means++}&&\multicolumn{2}{c}{$k$-means++}&&$imwk$&&$imwk$\\
\cline{2-3}
\cline{5-6}
\cline{12-13}
\cline{15-16}
&ARI&std&&ARI&std&&ARI&&ARI&&ARI&std&&ARI&std&&ARI&&ARI\\
\cline{1-20}
Australian CC&0.308&0.20&&0.504&0.00&&0.504&&0.524&&0.261&0.23&&0.334&0.08&&0.203&&0.504\\ 
Balance&0.132&0.04&&0.142&0.02&&0.172&&0.172&&0.130&0.04&&0.149&0.01&&0.172&&0.172\\ 
Breast cancer&0.839&0.00&&0.844&0.00&&0.850&&0.866&&0.826&0.01&&0.813&0.00&&0.828&&0.828\\ 
Car eval.&0.062&0.05&&0.219&0.00&&0.211&&0.224&&0.020&0.03&&0.102&0.00&&0.131&&0.224\\ 
Mines vs rocks&-0.004&0.00&&0.042&0.04&&0.058&&0.167&&0.001&0.00&&0.079&0.00&&0.049&&0.109\\ 
Ecoli&0.457&0.07&&0.550&0.09&&0.038&&0.745&&0.513&0.09&&0.647&0.09&&0.038&&0.449\\ 
Glass&0.185&0.04&&0.226&0.02&&0.279&&0.317&&0.180&0.06&&0.227&0.04&&0.223&&0.292\\ 
Heart&0.341&0.10&&0.321&0.06&&0.314&&0.385&&0.321&0.16&&0.396&0.18&&0.349&&0.442\\ 
Ionosphere&0.168&0.03&&0.209&0.00&&0.209&&0.209&&0.162&0.03&&0.209&0.00&&0.209&&0.209\\ 
Iris&0.680&0.09&&0.880&0.07&&0.904&&0.922&&0.591&0.07&&0.872&0.11&&0.886&&0.922\\ 
Lung cancer&0.115&0.07&&0.198&0.05&&0.210&&0.255&&0.067&0.07&&0.304&0.01&&0.307&&0.411\\ 
Musk&0.003&0.00&&0.010&0.00&&0.014&&0.029&&0.003&0.00&&0.013&0.00&&0.010&&0.082\\ 
Parkinsons&0.049&0.00&&0.000&0.00&&-0.069&&0.292&&-0.088&0.01&&0.021&0.00&&-0.027&&0.167\\ 
Soya&0.769&0.20&&0.819&0.10&&1.000&&1.000&&0.823&0.19&&0.861&0.07&&1.000&&1.000\\
Teaching a.&0.031&0.01&&0.030&0.02&&0.042&&0.075&&0.008&0.01&&0.022&0.01&&0.024&&0.074\\ 
Tic-tac-toe&0.016&0.02&&0.019&0.00&&0.019&&0.069&&0.014&0.01&&0.038&0.00&&0.019&&0.151\\ 
Wine&0.838&0.07&&0.836&0.02&&0.818&&0.850&&0.863&0.10&&0.828&0.06&&0.865&&0.852\\ 
Zoo&0.684&0.13&&0.943&0.00&&0.943&&0.962&&0.707&0.10&&0.911&0.06&&0.940&&0.940\\ 
\end{tabular}
\end{center}
\end{table}
\begin{table}[t]\footnotesize
\begin{center}
\caption{A comparison between the ARI values for $imwk$-means and its rescaled version, supplied with good values of the exponents $p_1$ and $p_2$, and the expected ARI given by $k$-means++. The presented results are the averages over 50 data sets for each of the configurations below, each normalised using z-scores.}
\label{Tab:Results_BestPossible_zstd}
\tabcolsep=0.08cm
\begin{tabular}{lccccccccccc}
&&&&\multicolumn{2}{c}{Rescaled}&&&&&\multicolumn{2}{c}{Rescaled}\\
&\multicolumn{2}{c}{\textit{k}-means++}&&\multicolumn{2}{c}{\textit{k}-means++}&&\multicolumn{2}{c}{\textit{imwk}-means}&&\multicolumn{2}{c}{\textit{imwk}-means}\\
\cline{2-3}
\cline{5-6}
\cline{8-9}
\cline{11-12}
&ARI&std&&ARI&std&&ARI&std&&ARI&std\\
1000x6-3&0.5060&0.214&&0.5677&0.217&&0.5888&0.218&&0.6617&0.189\\ 
1000x12-6&0.6336&0.173&&0.6739&0.155&&0.7412&0.173&&0.7996&0.129\\ 
1000x20-10&0.7708&0.112&&0.7968&0.085&&0.8981&0.076&&0.9376&0.053\\ 
\cline{1-12}
1000x6-3 +3NF&0.4550&0.211&&0.5360&0.209&&0.5793&0.214&&0.6595&0.195\\ 
1000x12-6 +6NF&0.5820&0.171&&0.6429&0.157&&0.7285&0.178&&0.8138&0.127\\ 
1000x20-10 +10NF&0.7233&0.125&&0.7710&0.087&&0.9024&0.067&&0.9400&0.040\\ 
\cline{1-12}
1000x6-3 +3NNF&0.4690&0.212&&0.5473&0.216&&0.5794&0.226&&0.6513&0.192\\ 
1000x12-6 +6NNF&0.5840&0.168&&0.6540&0.155&&0.7381&0.179&&0.7879&0.138\\ 
1000x20-10 +10NNF&0.7278&0.123&&0.7773&0.085&&0.8946&0.066&&0.9358&0.045\\ 
\cline{1-12}
1000x6-3 WCN&0.1645&0.095&&0.2596&0.151&&0.3635&0.180&&0.4144&0.171\\ 
1000x12-6 WCN&0.1835&0.070&&0.3952&0.151&&0.6128&0.169&&0.6256&0.167\\ 
1000x20-10 WCN&0.1735&0.048&&0.5593&0.107&&0.8386&0.068&&0.8405&0.072\\ 
\end{tabular}
\end{center}
\end{table}

We then conducted a new set of experiments aiming at determining whether there is a pattern for suitable parameters of the rescaled $imwk$-means. Given that we are interested in a general pattern, we performed such experiments only for synthetic data sets. Even if we were able to establish such a pattern for 18 real-world data sets, we would not be able to draw any realistic conclusion from it. In a set of experiments the first question one usually would ask is whether particular pairs of the exponent parameters $p_1$ and $p_2$ work well on average (over the 50 data sets for each configuration) for the rescaled $imwk$-means. Table \ref{Tab:Results_BestAvg_std} shows the results of these experiments for the data sets normalised using the range normalisation. The differences in ARI are not as large as before, but we can still see that the rescaled $imwk$-means is competitive and usually outperforms $imwk$-means. The same can be said when comparing the rescaled $k$-means++ and $k$-means++. Table \ref{Tab:Results_BestAvg_zstd} presents the results for data sets normalised using $z$-scores. In this case, the rescaled versions of the algorithms were still superior when compared to $imwk$-means and $k$-means++.
\begin{table}[t]\footnotesize
\begin{center}
\caption{A comparison between the ARI values for clusterings found by $imwk$-means and its rescaled version, supplied with parameters that work well on average, and the expected ARI given by $k$-means++. The presented results are the averages over 50 data sets for each of the configurations below, each normalised using the range normalisation.}
\label{Tab:Results_BestAvg_std}
\tabcolsep=0.05cm
\begin{tabular}{lcccccccccccccc}
&&&&\multicolumn{2}{c}{Rescaled}&&&&&&\multicolumn{4}{c}{Rescaled}\\
&\multicolumn{2}{c}{\textit{k}-means++}&&\multicolumn{2}{c}{\textit{k}-means++}&&\multicolumn{3}{c}{\textit{imwk}-means}&&\multicolumn{4}{c}{\textit{imwk}-means}\\
\cline{2-3}
\cline{5-6}
\cline{8-10}
\cline{12-15}
&ARI&std&&ARI&std&&ARI&std&$p$&&ARI&std&$p_1$&$p_2$\\
1000x6-3&0.5198&0.224&&0.5341&0.041&&0.5249&0.227&2.8&&0.5453&0.227&4.4&2.9\\
1000x12-6&0.6356&0.177&&0.6448&0.079&&0.6434&0.191&2.5&&0.6601&0.187&3.9&2.5\\ 
1000x20-10&0.7703&0.121&&0.7737&0.081&&0.8294&0.116&2.5&&0.8539&0.100&5.0&2.1\\ 
\cline{1-15}
1000x6-3 +3NF&0.0371&0.112&&0.4091&0.030&&0.4385&0.308&1.5&&0.4622&0.307&1.4&2.8\\ 
1000x12-6 +6NF&0.0997&0.112&&0.5970&0.078&&0.6820&0.190&1.6&&0.7152&0.195&1.7&2.4\\ 
1000x20-10 +10NF&0.1708&0.099&&0.7415&0.086&&0.7519&0.105&1.7&&0.8619&0.070&2.0&1.7\\ 
\cline{1-15}
1000x6-3 +3NNF&0.4748&0.225&&0.5121&0.057&&0.5236&0.225&2.4&&0.5341&0.219&4.9&2.6\\
1000x12-6 +6NNF&0.5852&0.172&&0.6266&0.079&&0.6539&0.175&2.0&&0.6518&0.183&4.7&2.4\\ 
1000x20-10+10NNF&0.7286&0.125&&0.7524&0.081&&0.8286&0.103&2.5&&0.8622&0.090&4.8&2.1\\ 
\cline{1-15}
1000x6-3 WCN&0.0597&0.059&&0.1916&0.009&&0.2524&0.208&1.6&&0.2798&0.215&4.7&1.5\\ 
1000x12-6 WCN&0.0920&0.062&&0.3635&0.042&&0.5677&0.191&1.5&&0.5791&0.192&1.5&2.9\\ 
1000x20-10 WCN&0.1049&0.032&&0.5296&0.050&&0.7594&0.116&1.6&&0.7690&0.107&5.0&1.7\\ 
\end{tabular}
\end{center}
\end{table}
\begin{table}[t]\footnotesize
\begin{center}
\caption{A comparison between the ARI values for clusterings found by $imwk$-means and its rescaled version, supplied with parameters that work well on average, and the expected ARI given by $k$-means++. The presented results are the averages over 50 data sets for each of the configurations below, each normalised using $z$-scores.}
\label{Tab:Results_BestAvg_zstd}
\tabcolsep=0.05cm
\begin{tabular}{lcccccccccccccc}
&&&&\multicolumn{2}{c}{Rescaled}&&&&&&\multicolumn{4}{c}{Rescaled}\\
&\multicolumn{2}{c}{\textit{k}-means++}&&\multicolumn{2}{c}{\textit{k}-means++}&&\multicolumn{3}{c}{\textit{imwk}-means}&&\multicolumn{4}{c}{\textit{imwk}-means}\\
\cline{2-3}
\cline{5-6}
\cline{8-10}
\cline{12-15}
&ARI&std&&ARI&std&&ARI&std&$p$&&ARI&std&$p_1$&$p_2$\\
1000x6-3&0.5060&0.214&&0.5326&0.037&&0.5286&0.221&2.7&&0.5406&0.229&3.9&2.6\\ 
1000x12-6&0.6336&0.173&&0.6488&0.079&&0.6533&0.190&3.0&&0.6463&0.184&4.3&2.5\\ 
1000x20-10&0.7708&0.112&&0.7714&0.080&&0.8299&0.104&2.1&&0.8489&0.110&3.8&2.1\\ 
\cline{1-15}
1000x6-3 +3NF&0.4550&0.211&&0.4805&0.069&&0.4771&0.200&3.8&&0.5143&0.239&3.1&3.1\\ 
1000x12-6 +6NF&0.5820&0.171&&0.6091&0.080&&0.6281&0.193&2.6&&0.6496&0.205&5.0&3.0\\ 
1000x20-10 +10NF&0.7233&0.125&&0.7447&0.080&&0.8221&0.110&2.5&&0.8567&0.104&3.2&2.0\\ 
\cline{1-15}
1000x6-3 +3NNF&0.4690&0.212&&0.5057&0.047&&0.5072&0.232&2.4&&0.5193&0.229&4.9&2.7\\ 
1000x12-6 +6NNF&0.5840&0.168&&0.6272&0.079&&0.6577&0.186&1.8&&0.6430&0.186&4.8&2.4\\ 
1000x20-10 +10NNF&0.7278&0.123&&0.7508&0.081&&0.8282&0.084&2.4&&0.8524&0.096&4.1&2.2\\ 
\cline{1-15}
1000x6-3 WCN&0.1645&0.095&&0.2082&0.014&&0.2466&0.189&2.0&&0.2672&0.154&3.4&2.6\\
1000x12-6 WCN&0.1835&0.070&&0.3535&0.038&&0.5199&0.204&1.6&&0.5529&0.209&1.4&2.2\\ 
1000x20-10 WCN&0.1735&0.048&&0.5280&0.051&&0.7496&0.115&1.6&&0.7682&0.143&1.4&2.1\\ 
\end{tabular}
\end{center}
\end{table}

Given the difficulty of finding clear patterns of good values for $p_1$ and $p_2$ for the rescaled $imwk$-means, we generated various figures showing the average ARI per pair $(p_1, p_2)$ for each data set considered in our simulations. For an easier comparison, we set to white each pixel representing a pair $(p_1, p_2)$ that did not outperform $k$-means++. Here, the experimented were carried out with the values of $p$ ranging from $1.1$ to $5.0$ (with a step of $0.1$).

\begin{figure}[t]
    \centering
    \begin{subfigure}[b]{0.3\textwidth}
        \includegraphics[width=\textwidth]{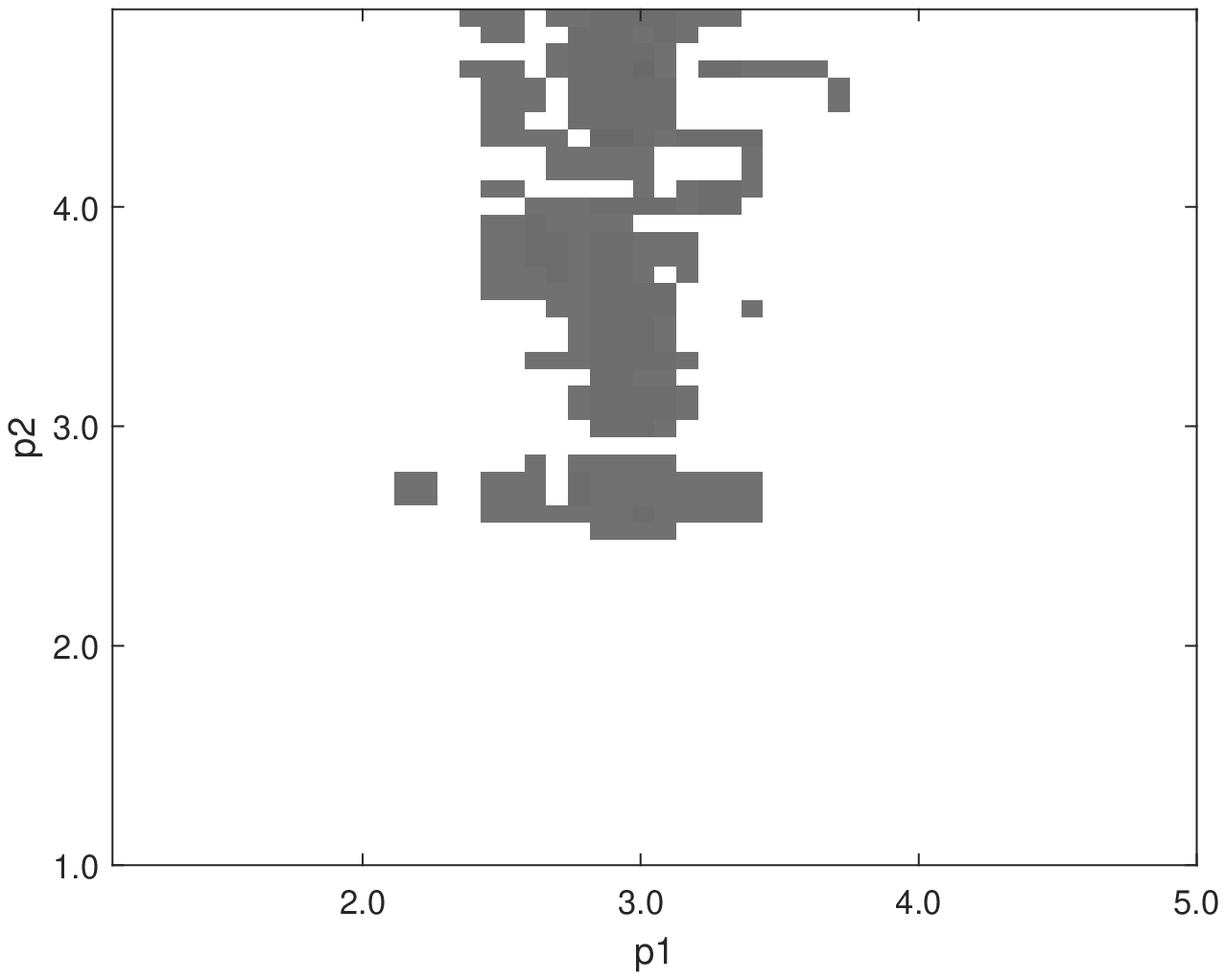}
        \caption{1000x6-3.}
        \label{Fig:6-3_std}
    \end{subfigure}
    \begin{subfigure}[b]{0.3\textwidth}
        \includegraphics[width=\textwidth]{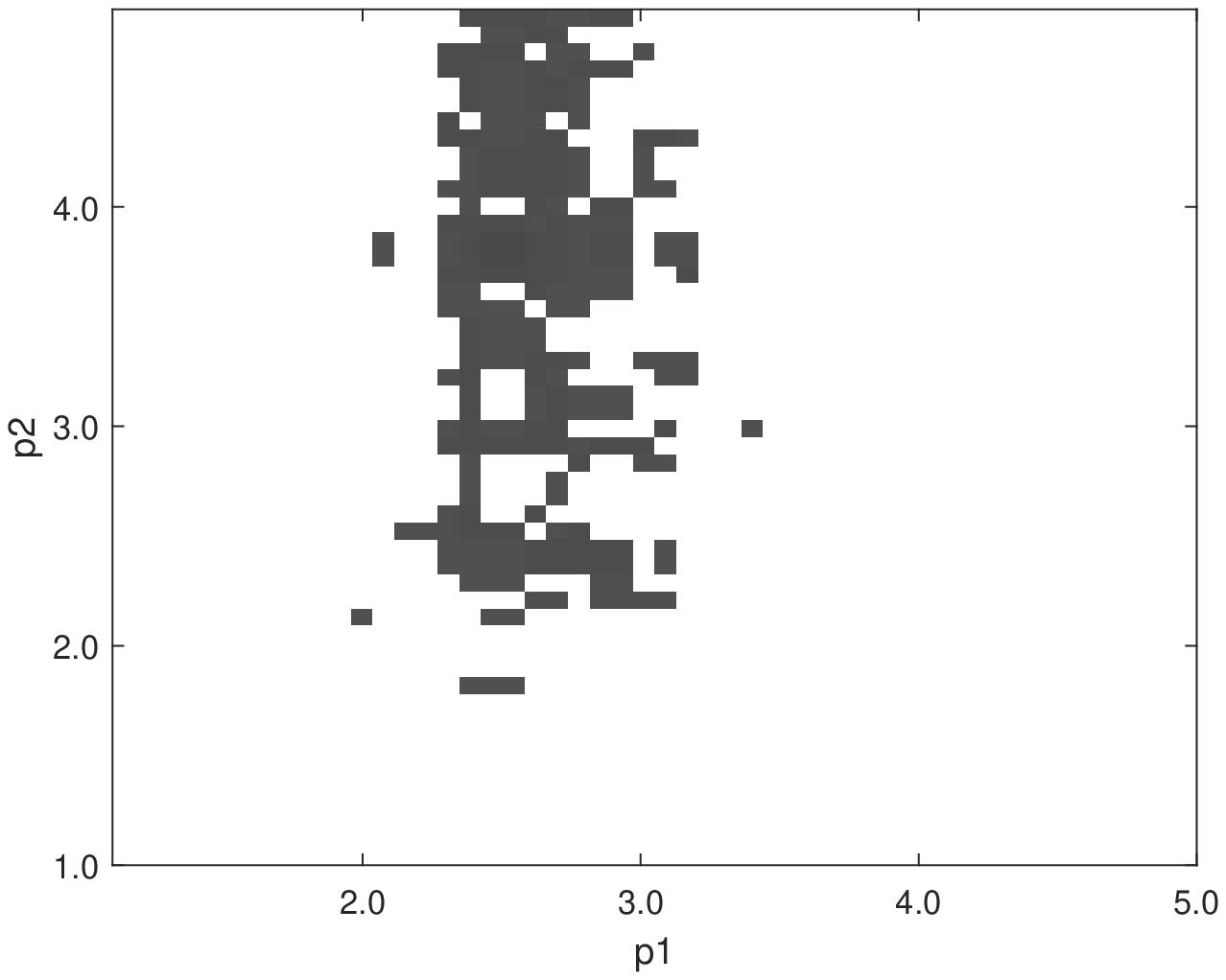}
        \caption{1000x12-6.}
        \label{Fig:12-6_std}
    \end{subfigure}
    \begin{subfigure}[b]{0.3\textwidth}
        \includegraphics[width=\textwidth]{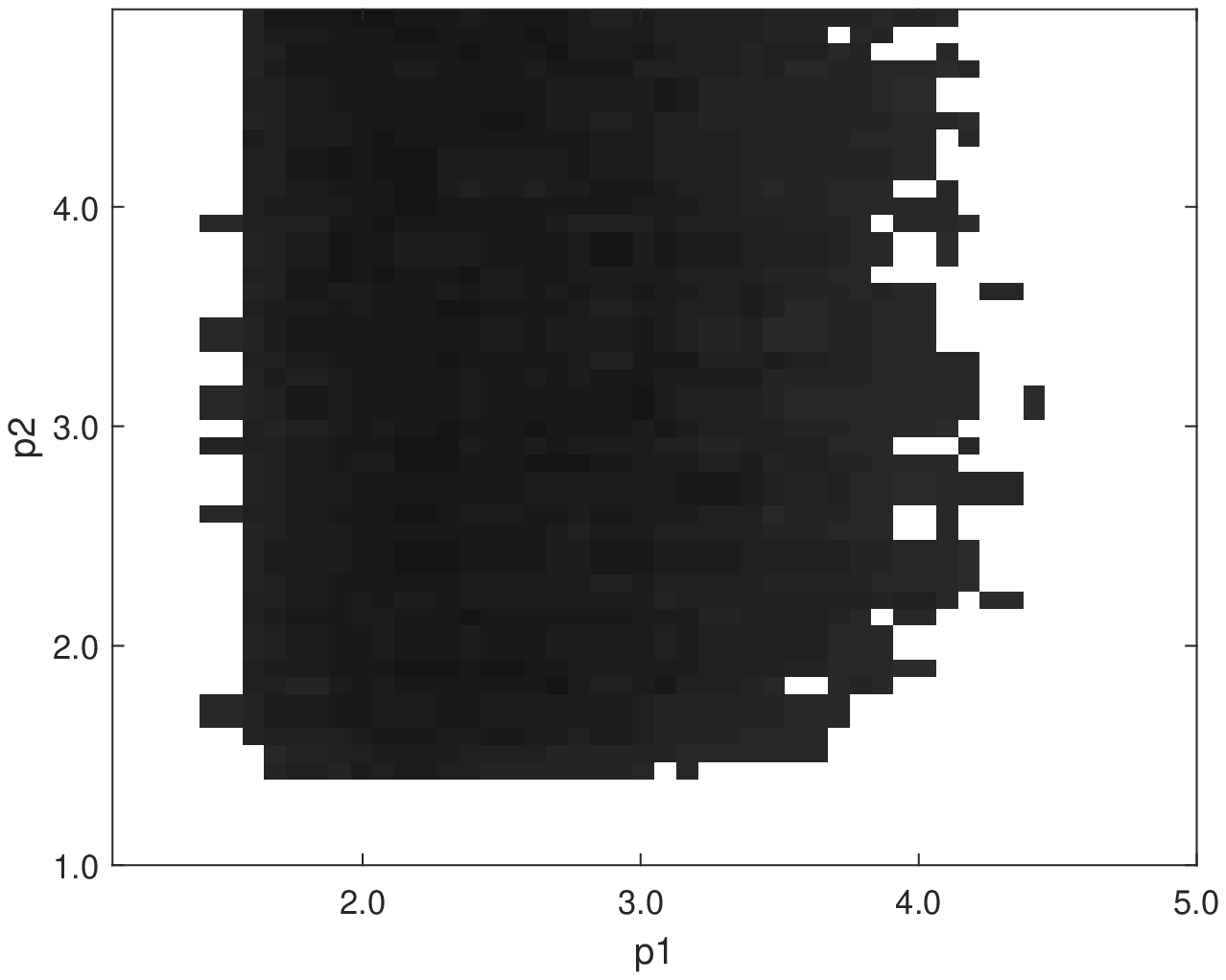}
        \caption{1000x20-10.}
        \label{Fig:20-10_std}
    \end{subfigure}
    \begin{subfigure}[b]{0.3\textwidth}
        \includegraphics[width=\textwidth]{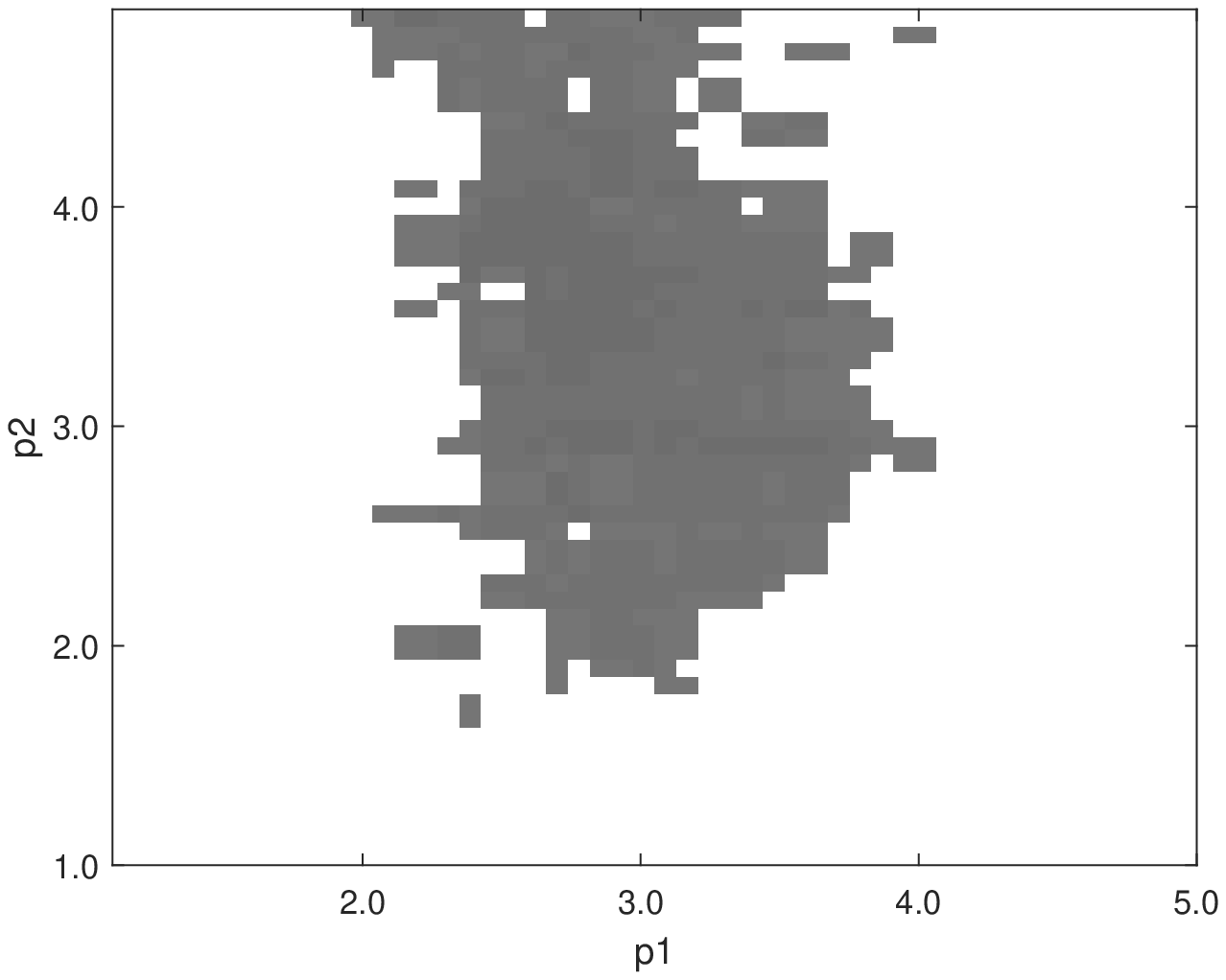}
        \caption{1000x6-3.}
        \label{Fig:6-3_zstd}
    \end{subfigure}
    \begin{subfigure}[b]{0.3\textwidth}
        \includegraphics[width=\textwidth]{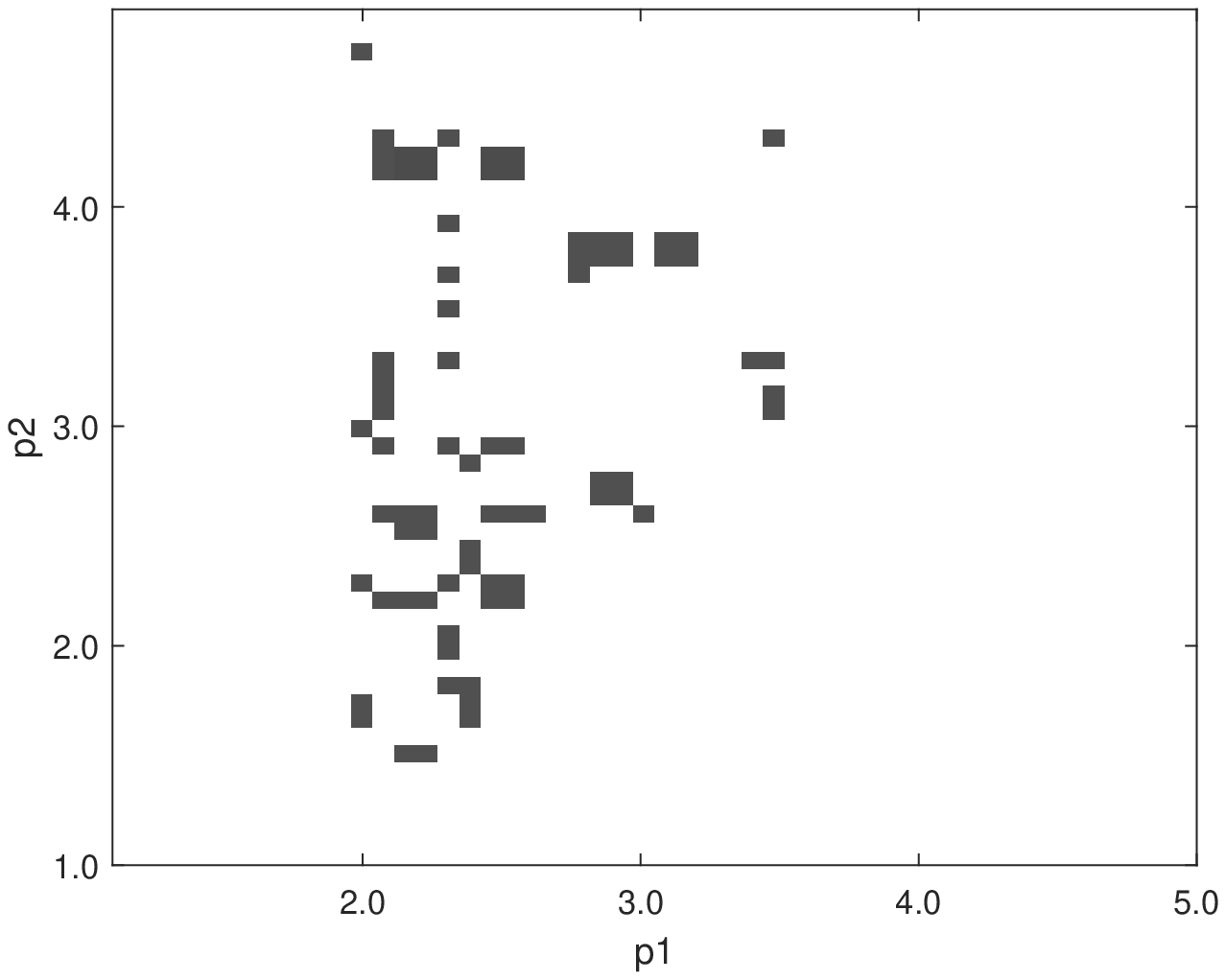}
        \caption{1000x12-6.}
        \label{Fig:12-6_zstd}
    \end{subfigure}
    \begin{subfigure}[b]{0.3\textwidth}
        \includegraphics[width=\textwidth]{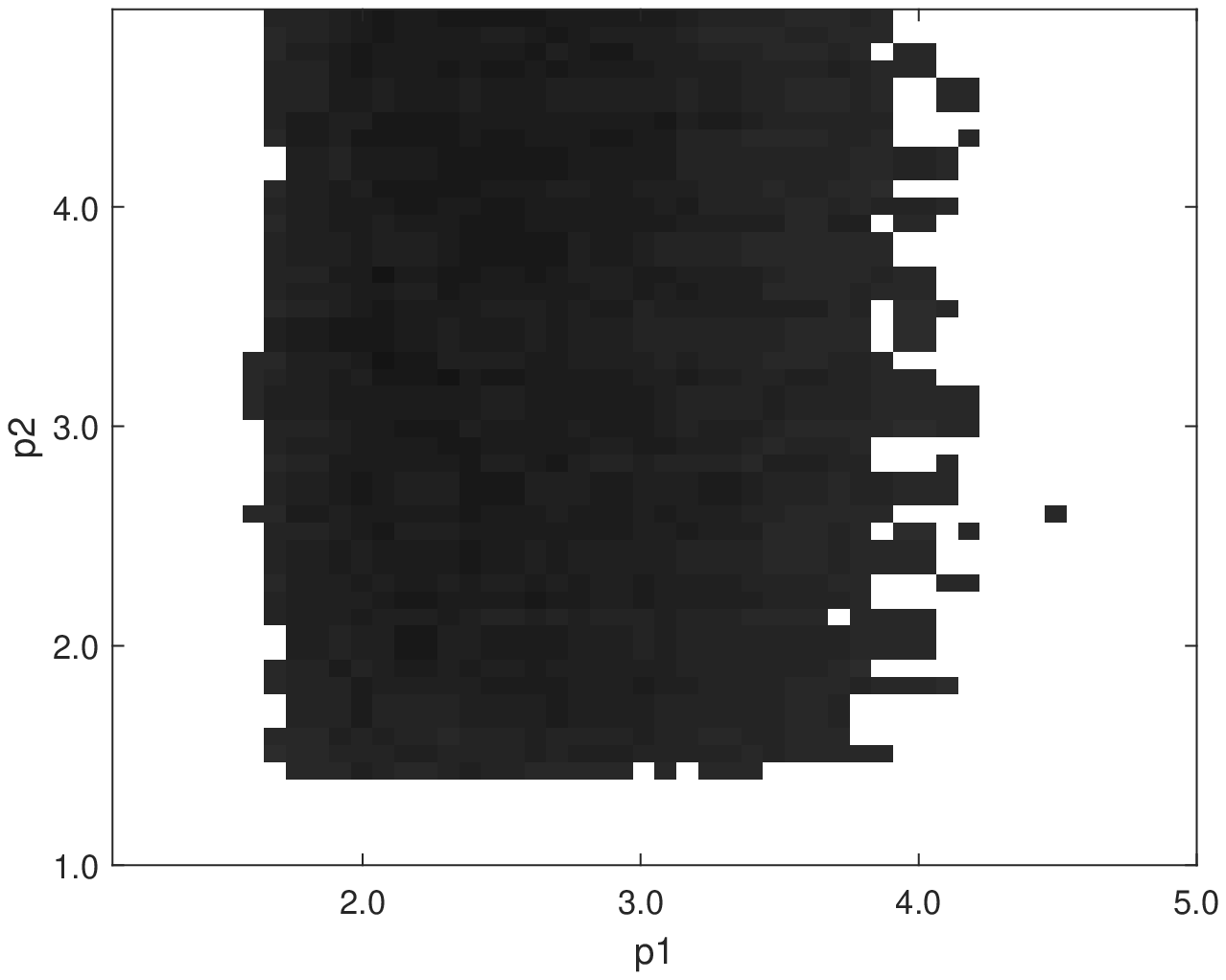}
        \caption{1000x20-10.}
        \label{Fig:20-10_zstd}
    \end{subfigure}
    \caption{Average ARIs provided by the rescaled $imwk$-means for each pair of the exponent parameters $p_1$ and $p_2$. No noise features were added to the data. White pixels represent pairs $(p_1, p_2)$ that did not outperform $k$-means++. We applied the range normalisation to the data sets in Figures \ref{Fig:6-3_std}, \ref{Fig:12-6_std} and \ref{Fig:20-10_std}. We applied $z$-scores to the data sets in Figures \ref{Fig:6-3_zstd}, \ref{Fig:12-6_zstd} and \ref{Fig:20-10_zstd}.}
     \label{Fig:NoNoise}
\end{figure}

We begin our analysis describing the results of the experiments with noise-free data. Figure \ref{Fig:NoNoise} shows the ARI results for the experiments with these data sets normalised using the range normalisation and $z$-scores. The presented results indicate that in both cases (range normalisation and $z$-scores) there is a limited number of pairs $(p_1, p_2)$ that led to high values of ARI for the data configurations 1000x6-3 and 1000x12-6. However, when the number of features and clusters increased (and by consequence the difficulty of producing a good clustering), we can observe that the majority of the exponent parameters $(p_1, p_2)$ led to high values of ARI. For instance, for the data configuration 1000x20-10 nearly all pairs, $(p_1, p_2)$, such that $p_1$ is located in the interval $[2,4]$ led to a high value of ARI.

\begin{figure}[t]
    \centering
    \begin{subfigure}[b]{0.3\textwidth}
        \includegraphics[width=\textwidth]{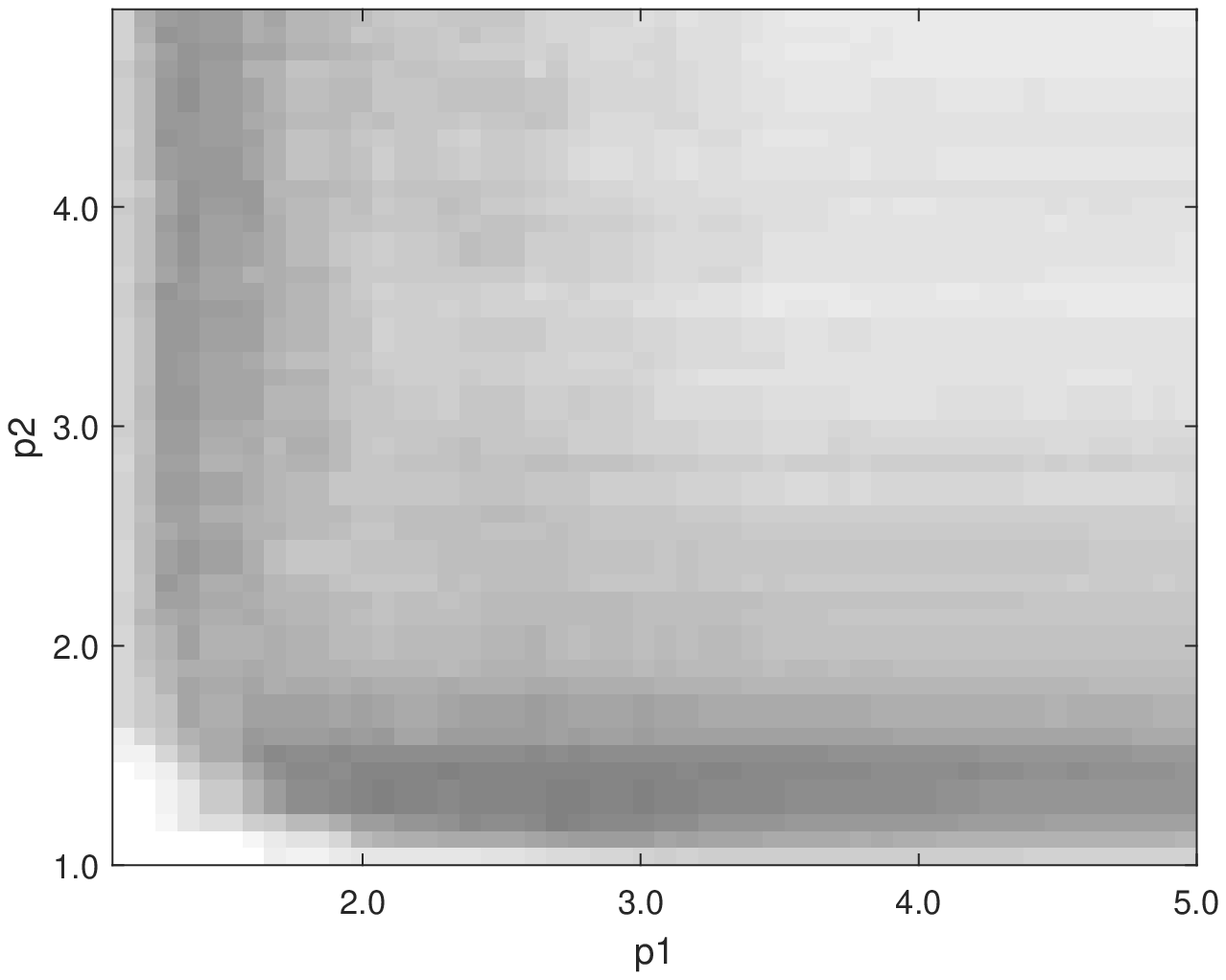}
        \caption{1000x6-3 +3NF.}
        \label{Fig:6-3NF_std}
    \end{subfigure}
    \begin{subfigure}[b]{0.3\textwidth}
        \includegraphics[width=\textwidth]{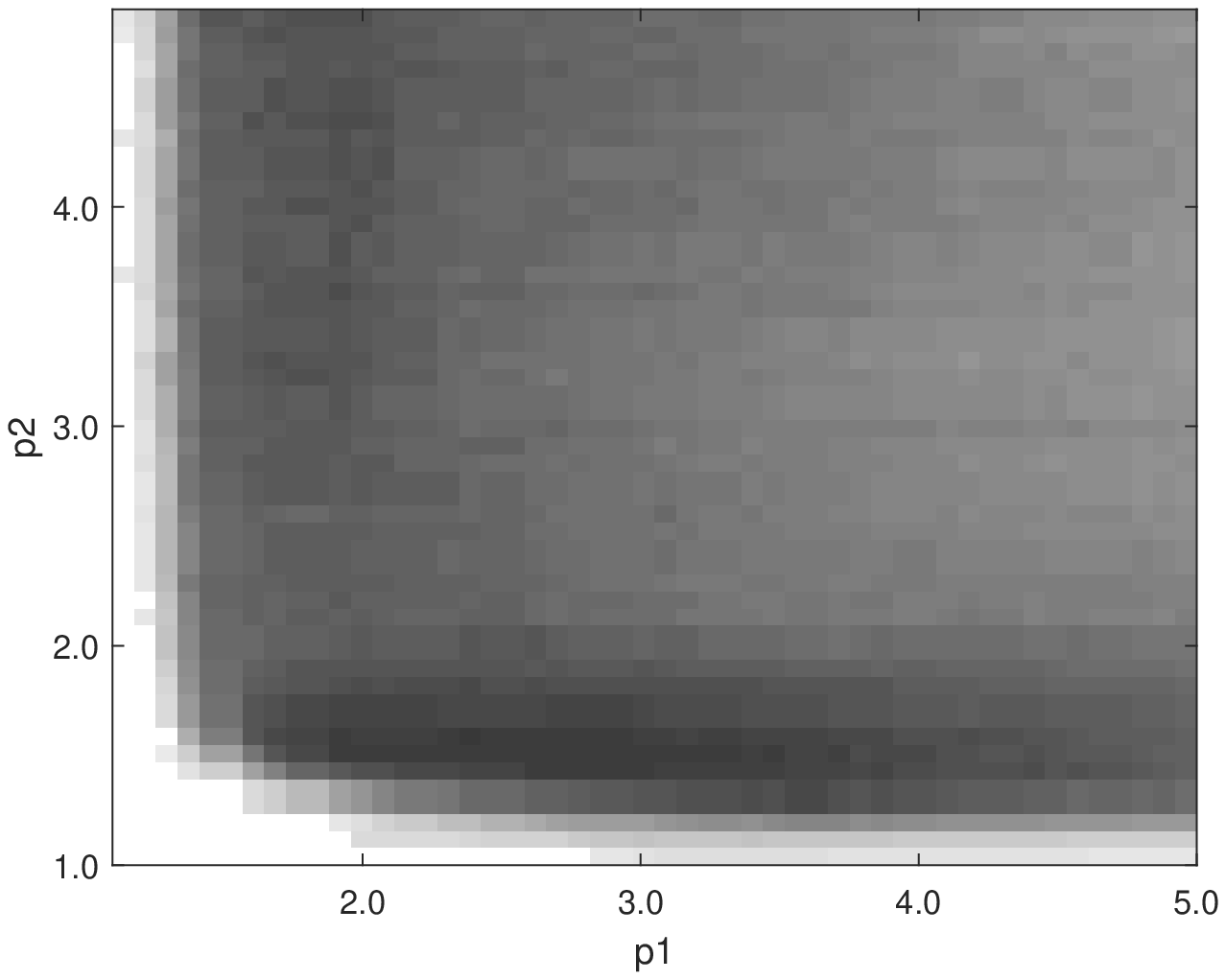}
        \caption{1000x12-6 +6NF.}
        \label{Fig:12-6NF_std}
    \end{subfigure}
    \begin{subfigure}[b]{0.3\textwidth}
        \includegraphics[width=\textwidth]{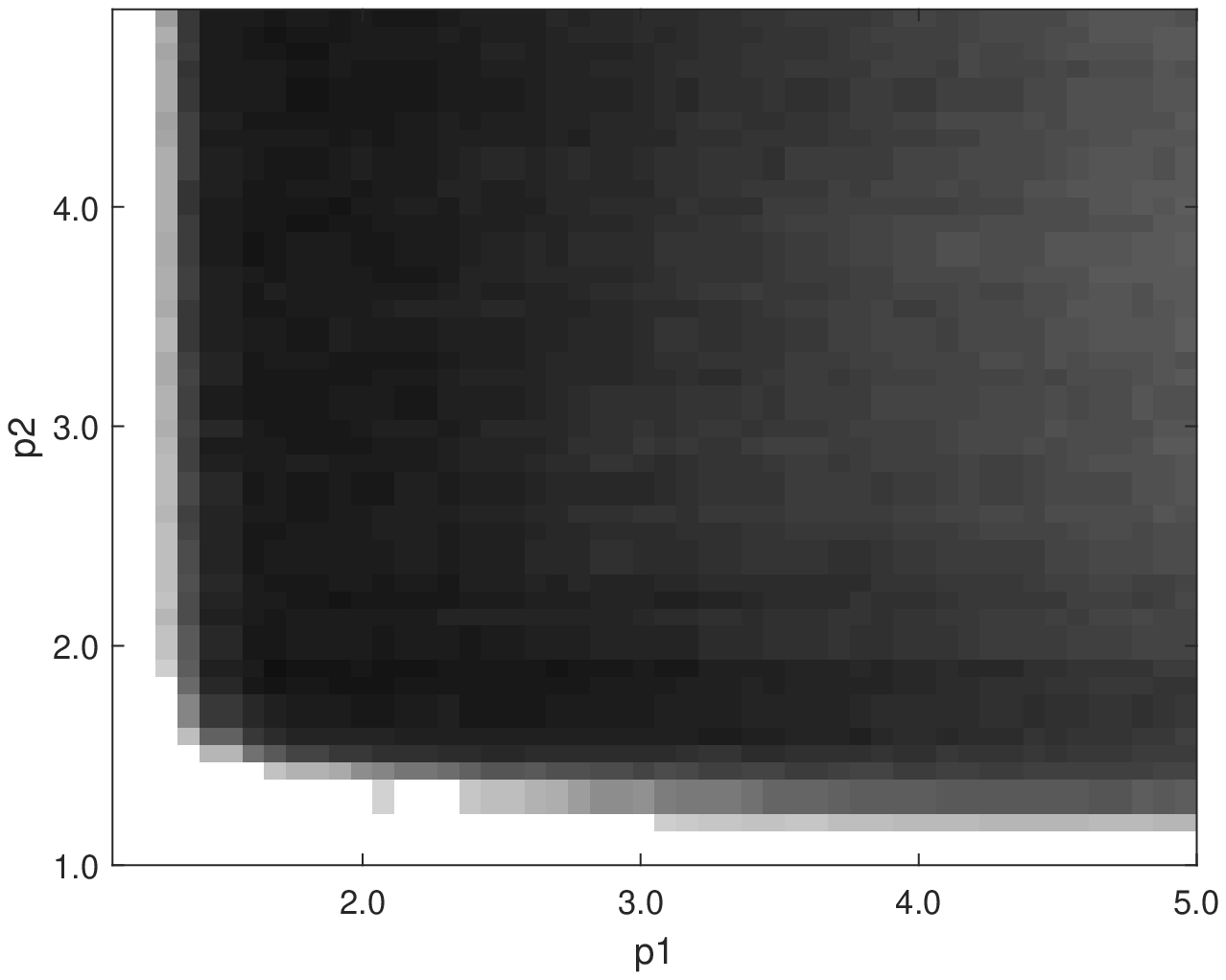}
        \caption{1000x20-10 +10NF.}
        \label{Fig:20-10NF_std}
    \end{subfigure}
    \begin{subfigure}[b]{0.3\textwidth}
        \includegraphics[width=\textwidth]{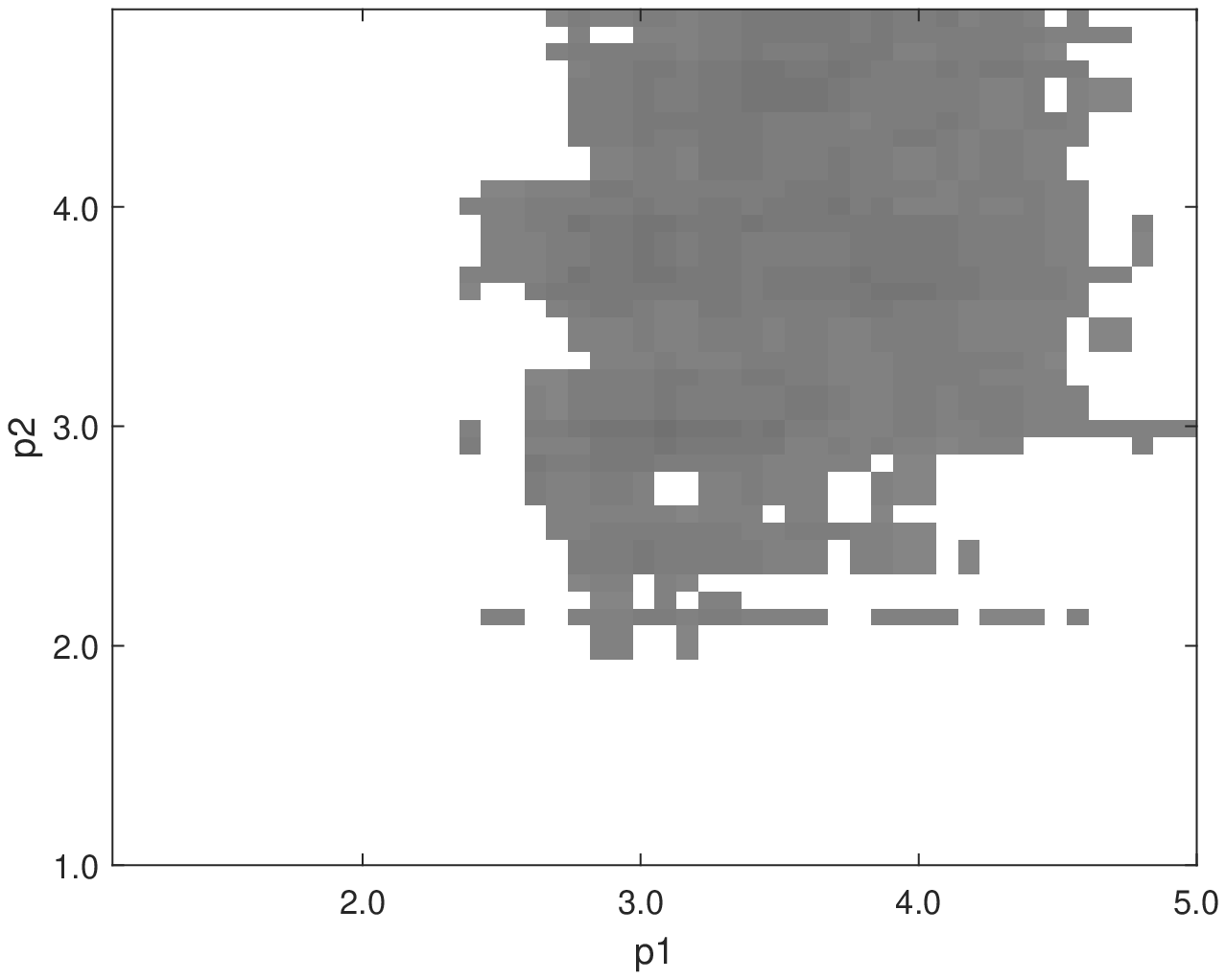}
        \caption{1000x6-3 +3NF.}
        \label{Fig:6-3NF_zstd}
    \end{subfigure}
    \begin{subfigure}[b]{0.3\textwidth}
        \includegraphics[width=\textwidth]{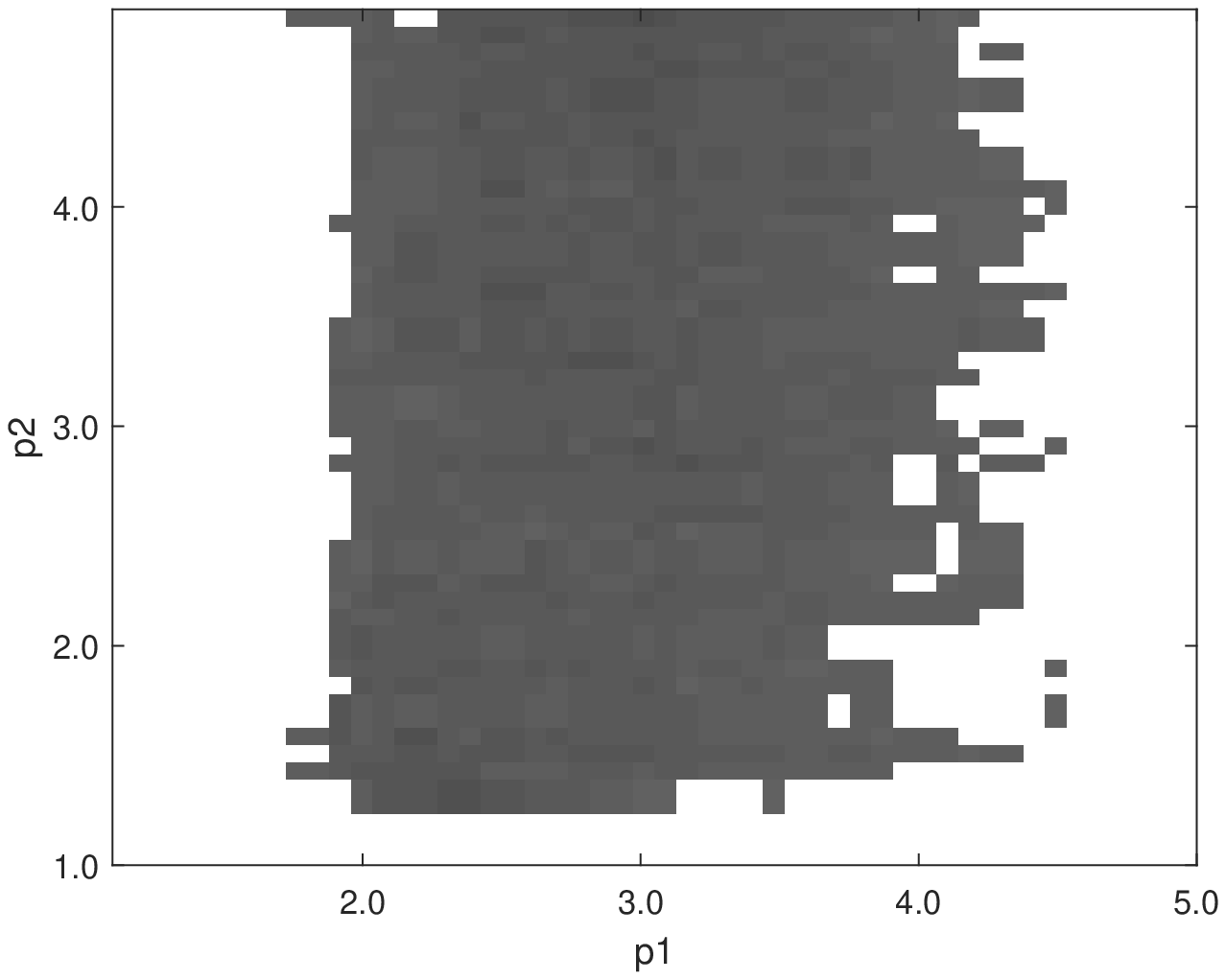}
        \caption{1000x12-6 +6NF.}
        \label{Fig:12-6NF_zstd}
    \end{subfigure}
    \begin{subfigure}[b]{0.3\textwidth}
        \includegraphics[width=\textwidth]{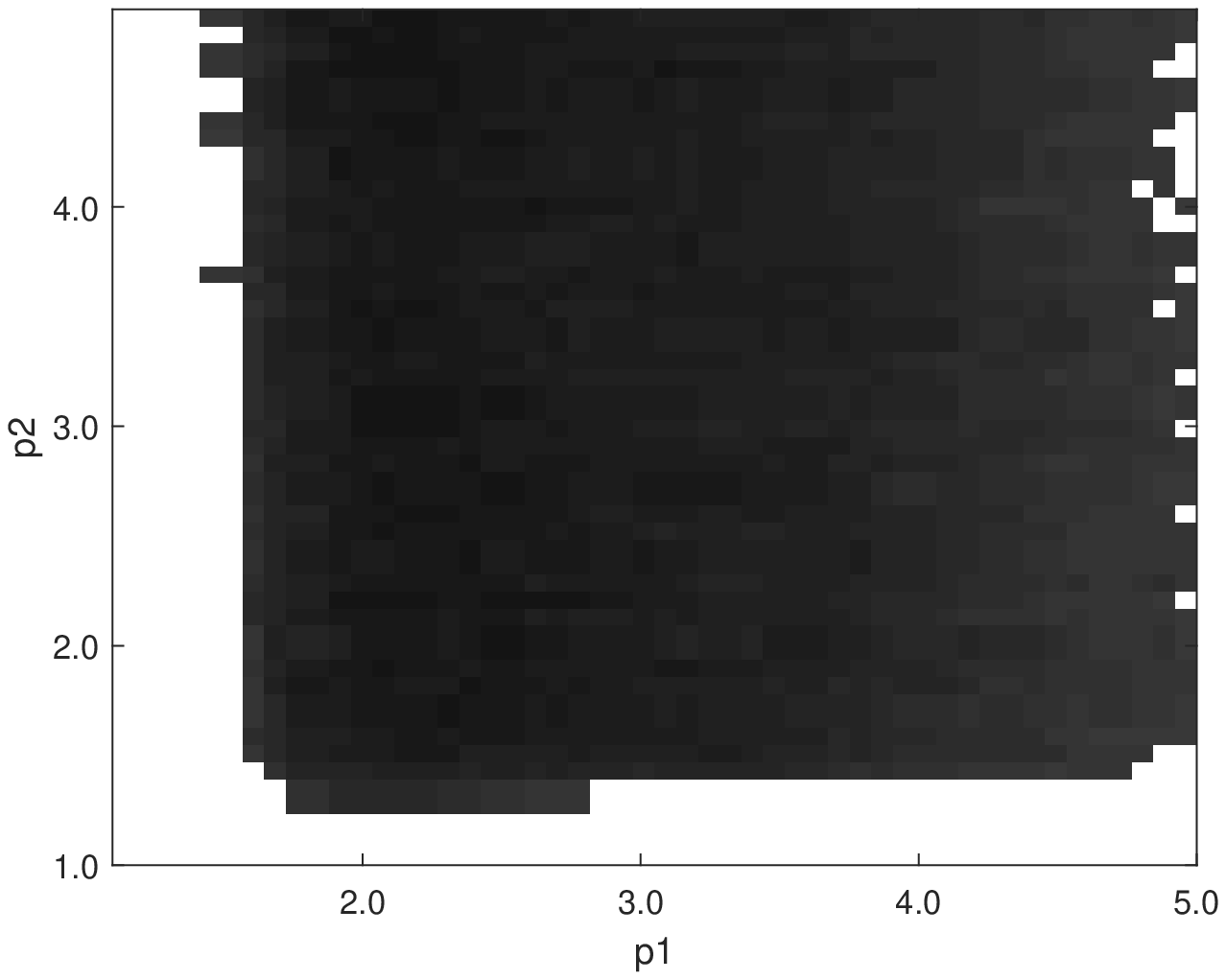}
        \caption{1000x20-10 +10NF.}
        \label{Fig:20-10NF_zstd}
    \end{subfigure}
    \caption{Average ARIs provided by the rescaled $imwk$-means for each pair of the exponent parameters $p_1$ and $p_2$. White pixels represent pairs $(p_1, p_2)$ that did not outperform $k$-means++. The noise features (NF) were composed of uniformly random variables. We applied the range normalisation to the data sets in Figures \ref{Fig:6-3NF_std}, \ref{Fig:12-6NF_std} and \ref{Fig:20-10NF_std}. We applied $z$-scores to the data sets in Figures \ref{Fig:6-3NF_zstd}, \ref{Fig:12-6NF_zstd} and \ref{Fig:20-10NF_zstd}.}
     \label{Fig:NF}
\end{figure}
\begin{figure}[t]
    \centering
    \begin{subfigure}[b]{0.3\textwidth}
        \includegraphics[width=\textwidth]{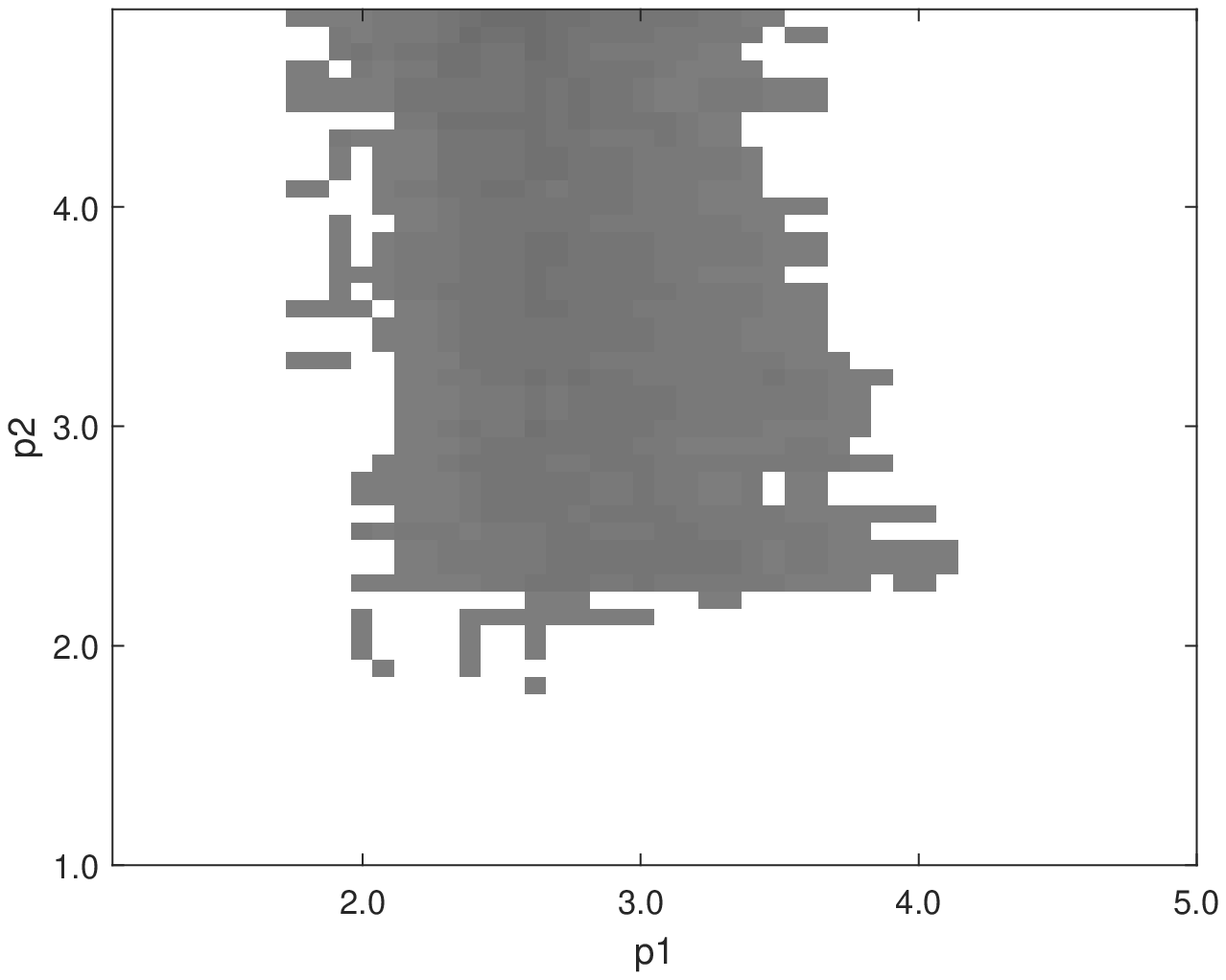}
        \caption{1000x6-3 +3NNF.}
        \label{Fig:6-3NNF_std}
    \end{subfigure}
    \begin{subfigure}[b]{0.3\textwidth}
        \includegraphics[width=\textwidth]{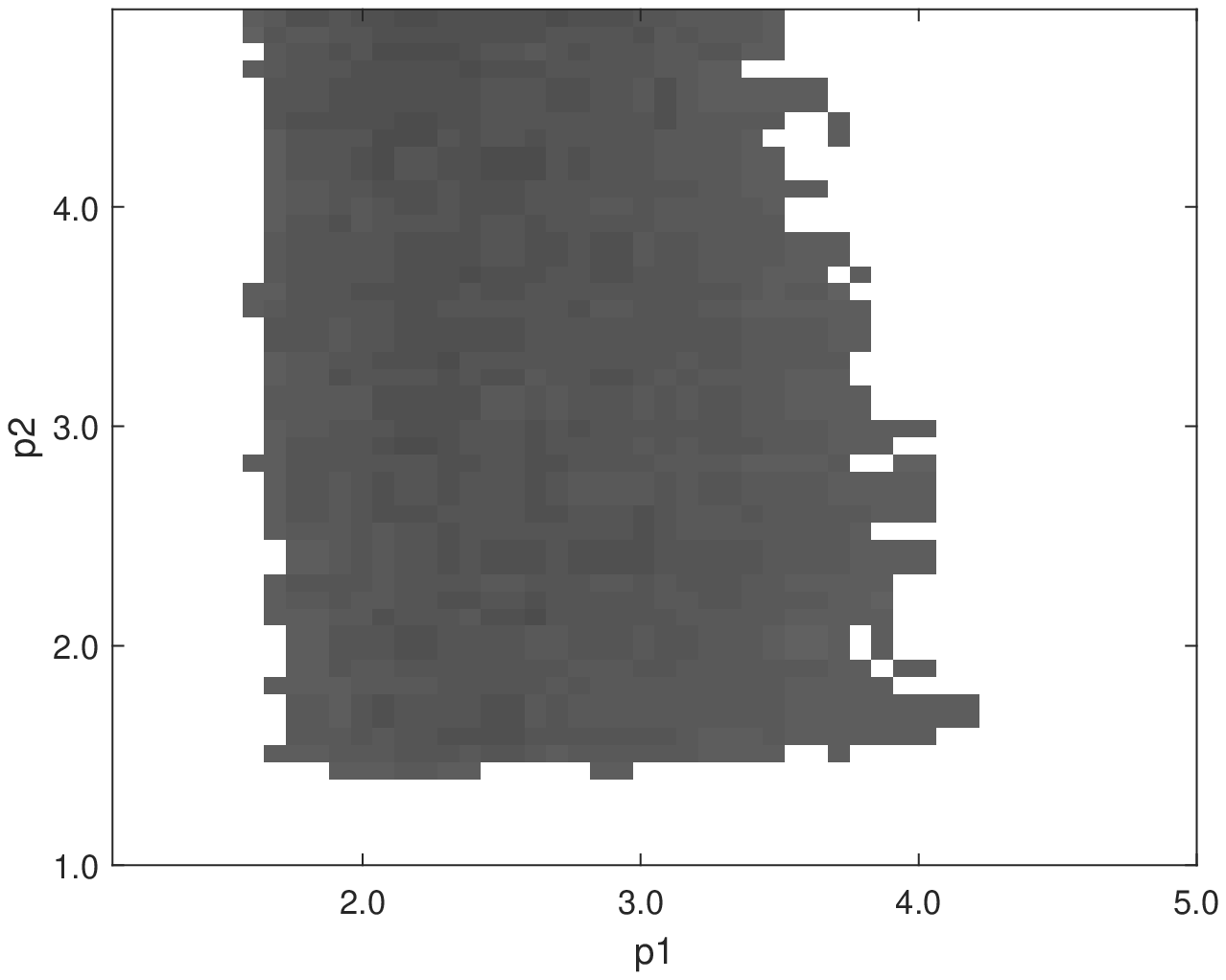}
        \caption{1000x12-6 +6NNF.}
        \label{Fig:12-6NNF_std}
    \end{subfigure}
    \begin{subfigure}[b]{0.3\textwidth}
        \includegraphics[width=\textwidth]{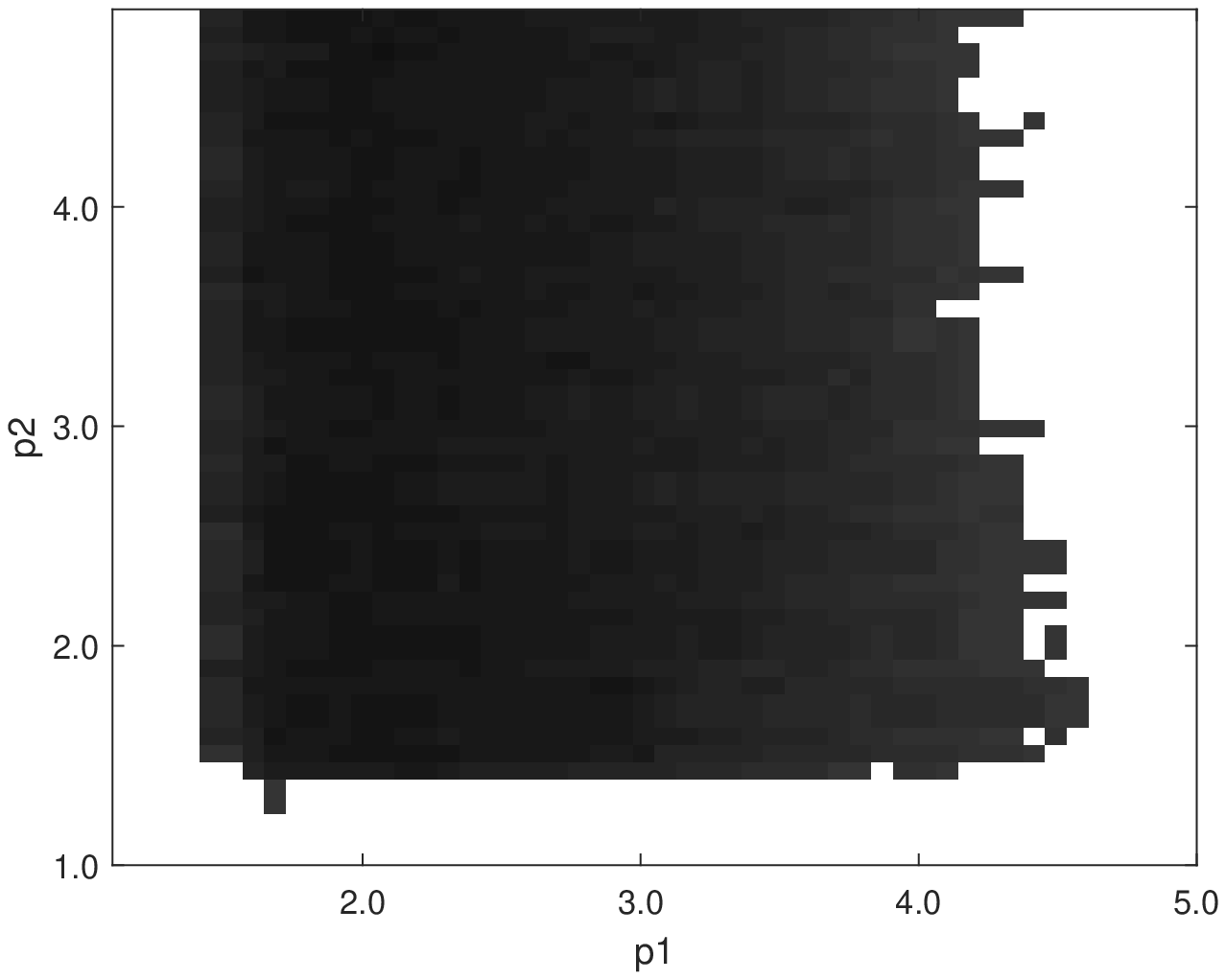}
        \caption{1000x20-10 +10NNF.}
        \label{Fig:20-10NNF_std}
    \end{subfigure}
    \begin{subfigure}[b]{0.3\textwidth}
        \includegraphics[width=\textwidth]{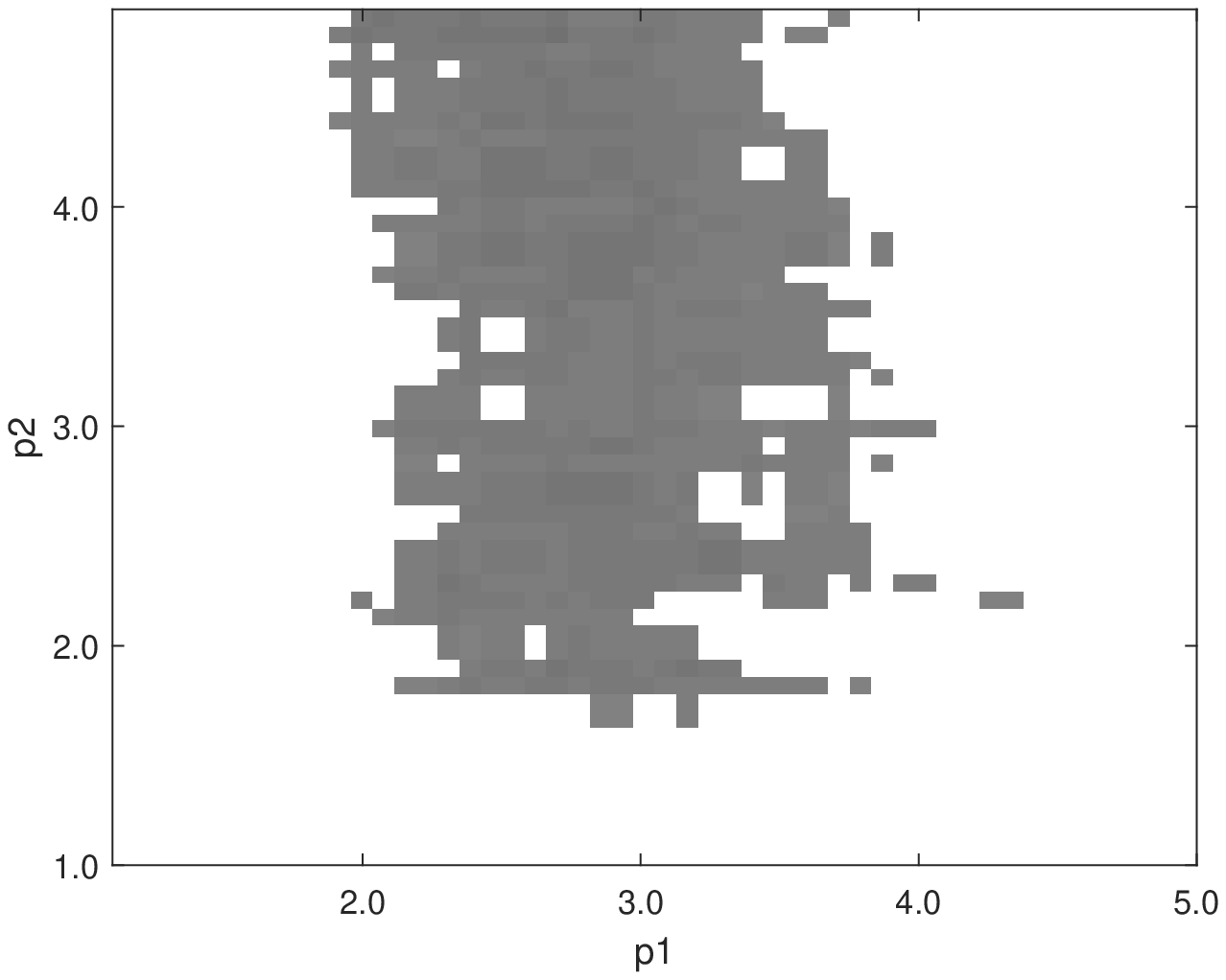}
        \caption{1000x6-3 +3NNF.}
        \label{Fig:6-3NNF_zstd}
    \end{subfigure}
    \begin{subfigure}[b]{0.3\textwidth}
        \includegraphics[width=\textwidth]{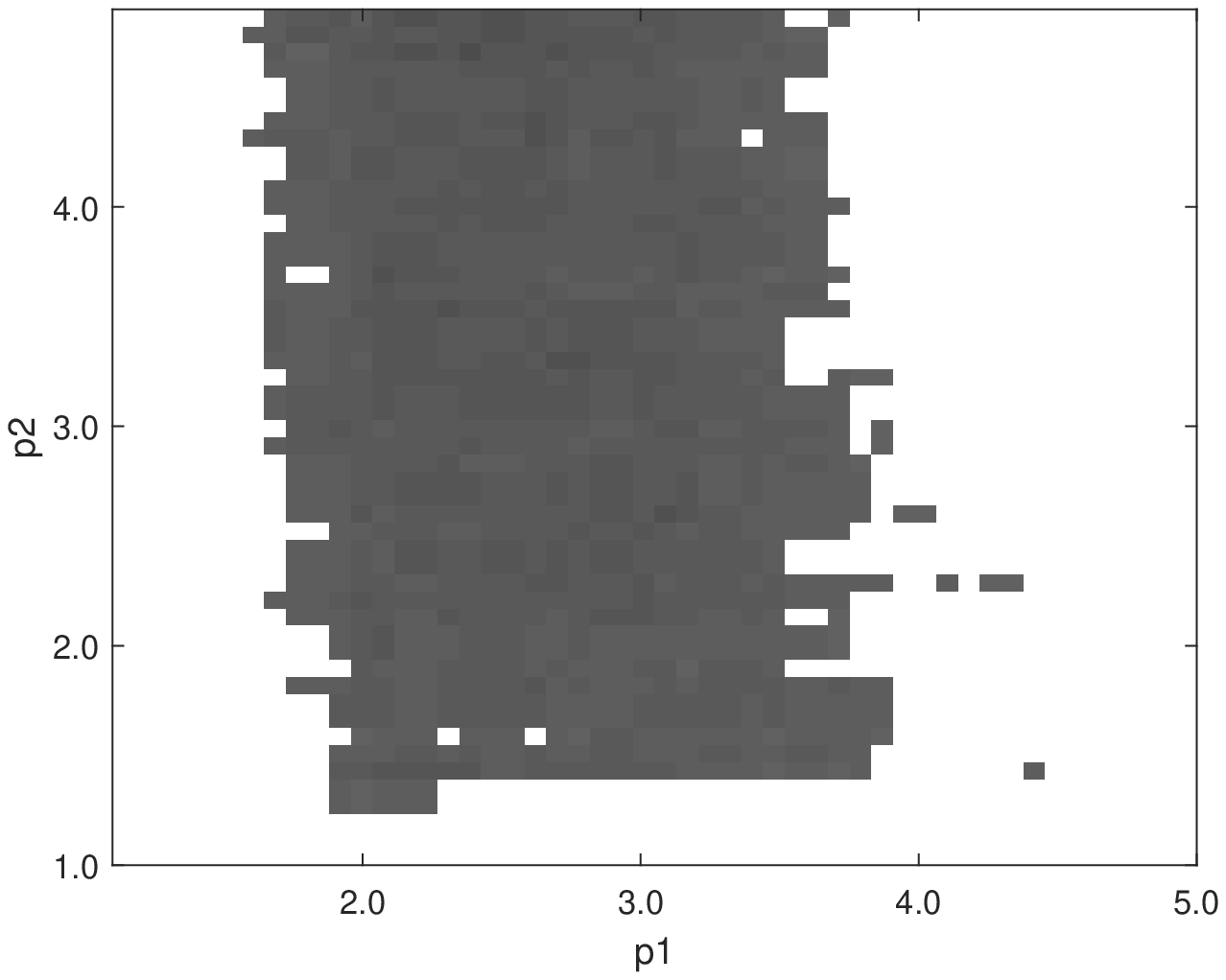}
        \caption{1000x12-6 +6NNF.}
        \label{Fig:12-6NNF_zstd}
    \end{subfigure}
    \begin{subfigure}[b]{0.3\textwidth}
        \includegraphics[width=\textwidth]{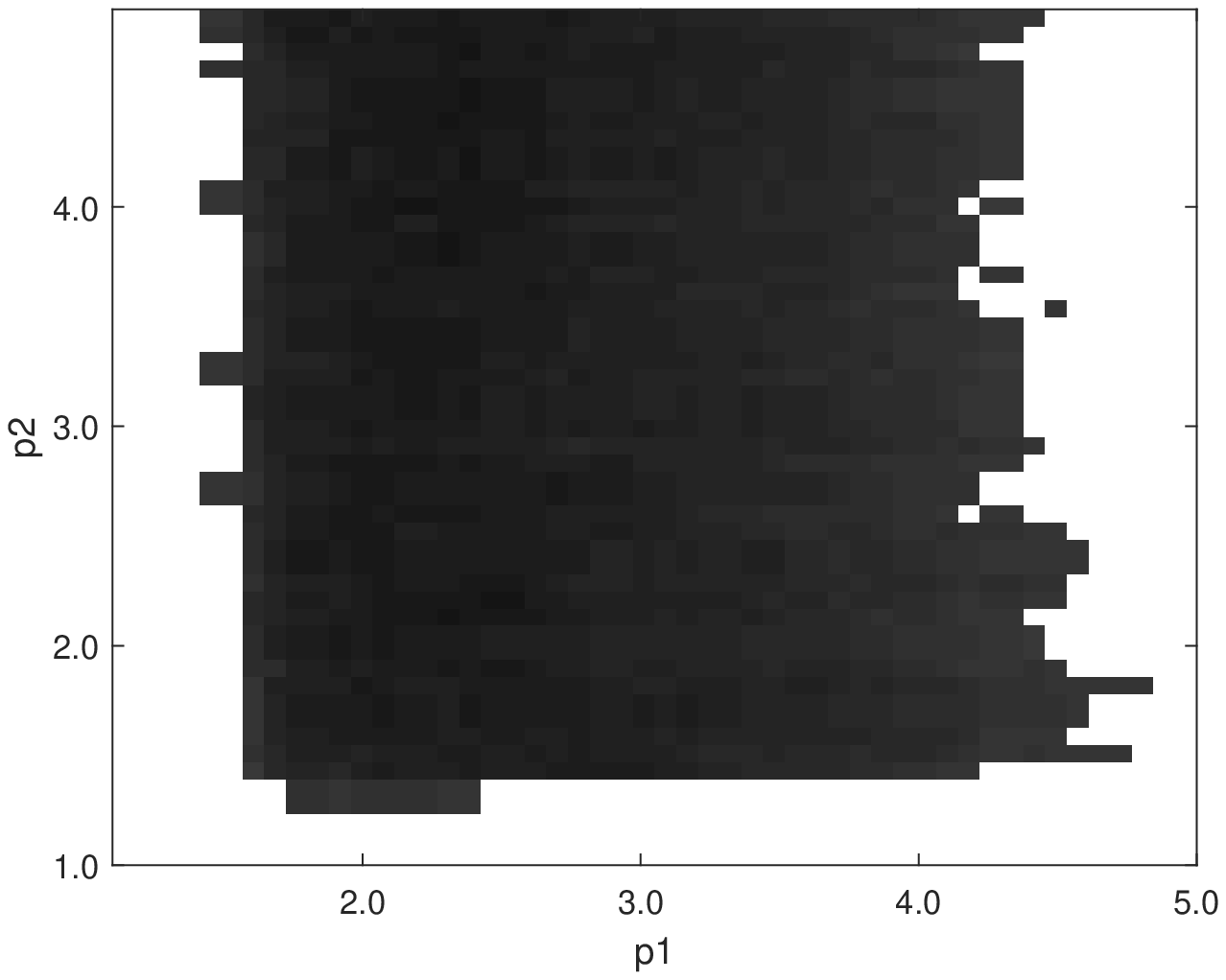}
        \caption{1000x20-10 +10NNF.}
        \label{Fig:20-10NNF_zstd}
    \end{subfigure}
    \caption{Average ARIs provided by the rescaled $imwk$-means for each pair of the exponent parameters $p_1$ and $p_2$. White pixels represent pairs $(p_1, p_2)$ that did not outperform $k$-means++. The noise features were composed of normally distributed random variables (NNF). We applied the range  normalisation to the data sets in Figures \ref{Fig:6-3NF_std}, \ref{Fig:12-6NF_std} and \ref{Fig:20-10NF_std}. We applied $z$-scores to the data sets in Figures \ref{Fig:6-3NF_zstd}, \ref{Fig:12-6NF_zstd} and \ref{Fig:20-10NF_zstd}.}
     \label{Fig:NNF}
\end{figure}
Figure \ref{Fig:NF} shows an even more favourably pattern for the proposed rescaled $imwk$-means. These experiments concern data sets to which we have added noise features containing uniformly random variables. When the data sets have been normalised used the range normalisation, nearly all possible pairs $(p_1, p_2)$ led to a considerably high value of ARI. When the data sets have been normalised with $z$-scores, we can clearly see that the more complex a data set, the larger the pool of pairs $(p_1, p_2)$ producing a high value of ARI is. In the case of the data sets under the configuration 1000x20-10 +10NF, nearly all pairs $(p_1, p_2)$ led to a high value of ARI. 

Figure \ref{Fig:NNF} shows a similar pattern. It illustrates the results of the experiments conducted with data sets to which we have added noise features containing Gaussian random values. Still, the more complex the data set, the greater the number of pairs $(p_1, p_2)$ producing high values of ARI is. We can also see that the difference between the ARI values generated by the rescaled $imwk$-means and $k$-means++ becomes larger. Again, in the case of the data configuration 1000x20-10 +10NNF, nearly all pairs of the exponent parameters led to a high value of ARI. 

Overall, the results of our experiments clearly indicate that rescaling a data set using our method improves cluster recovery. This is hardly surprising given that our feature rescaling factors minimise the within-cluster sum of distances (see Section \ref{Sec:NewAlg}). Thus, our method can certainly be used during the data preprocessing stage of any clustering algorithm based on distance measures. Of course, one could ask why our method improves the cluster recovery of methods capable of applying feature weighting (eg. $imwk$-means). This happens because such methods start with a suboptimal set of weights, sometimes containing even random values (for details see the recent surveys \cite{de2016survey,deng2016survey,kriegel2009clustering,kriegel2012subspace} and references there in). Usually, these weights are then improved at each iteration. Our rescaling method leads to more compact clusters, so the effect of the final set of weights (i.e. the most improved) produced by $imwk$-means can be experienced from the first iteration. 

\section{Conclusion}
\label{Sec:Conclusion}

Feature rescaling is a crucial part of the data preprocessing step in clustering. Typically original features describe entities located at different scales. Rescaling aims at balancing such features so that none of them overpowers the others in the objective function of the selected clustering algorithm. Here, we highlight that feature rescaling should in fact favour more meaningful features rather than simply put all of them on the same scale.

We introduced a data rescaling method based on the $imwk$-means algorithm. This algorithm applies to a normalised data set and produces a set of cluster-based feature weights. It is indeed intuitive that some features should be more relevant to certain clusters than the others. We showed how the feature weights can be used to account for the degree of relevance of any given feature at a particular cluster. These cluster-based weights are then used as feature rescaling factors. Our rescaling approach is quite different from the classical ones because a given feature will be rescaled using $k$ different factors, where $k$ is the number of clusters considered. The presented approach can be used in the data preprocessing step of any distance-based clustering algorithm such as $k$-means, $k$-means++, $imwk$-means, etc. 

Our approach works well because the feature weights minimise our clustering criteria (\ref{Eq:imwk}). Rescaling a data set using these weights as feature rescaling factors leads to more compact clusters, which are by consequence easier to be identified by a clustering algorithm. We demonstrated that our data preprocessing method generally produces a better cluster recovery than the standard normalisation procedures in a series of simulations involving synthetic and real-world data sets. The simulations on synthetic data sets were carried out with three types of noise features. First, we considered noise features containing uniformly random values. Second, we considered noise features containing Gaussian random values. Third, we considered features containing within cluster noise. 

We would like to suggest three directions for future research. They are as follows: (i) it would be interesting to investigate how the proposed feature rescaling method could be used as a data preprocessing step in the framework of supervised and semi-supervised machine learning approaches; (ii) we intend to investigate how our method behaves under other noise conditions; (iii) we are also interested in extending our method so that it would become capable to deal with data sets containing high numbers of features.

\bibliography{iMWK_ClusteringReScale}
\end{document}